\pdfoutput=1
\documentclass{article}

\PassOptionsToPackage{numbers, compress}{natbib}
\usepackage[main, final]{neurips_2026}

\usepackage[utf8]{inputenc}
\usepackage[T1]{fontenc}
\usepackage[
  colorlinks=true,
  linkcolor=blue2,
  citecolor=blue2,
  urlcolor=blue2
]{hyperref}
\hypersetup{pdftitle={Transformers Converge to Invariant Algorithmic Cores}, pdfauthor={Joshua S. Schiffman}}
\usepackage{url}
\usepackage{booktabs}
\usepackage{amsfonts}
\usepackage{amsmath}
\usepackage{amssymb}
\usepackage{mathtools}
\usepackage{bm}
\usepackage{nicefrac}
\usepackage{microtype}
\usepackage{xcolor}
\usepackage{graphicx}
\usepackage{float}
\usepackage{multirow}
\usepackage{array}
\usepackage{subcaption}
\usepackage{tcolorbox}
\tcbuselibrary{skins,breakable}
\usepackage{multicol}

\usepackage{mdframed}
\usepackage{xcolor}
\definecolor{blue2}{rgb}{0.0, 0.2, 0.6}
\definecolor{steelblue}{rgb}{0.27, 0.51, 0.71}
\definecolor{steelblue4}{HTML}{36648b}
\definecolor{iscolor}{RGB}{67,133,203}
\definecolor{arecolor}{RGB}{214,73,51}
\definecolor{wascolor}{RGB}{126,87,194}
\definecolor{werecolor}{RGB}{56,142,60}

\newcommand{\R}{\mathbb{R}}

\newcommand{\figpanel}[2]{\hyperref[#1]{Fig.~\ref*{#1}#2}}
\newcommand{\tabl}[1]{\hyperref[#1]{Table~\ref*{#1}}}
\newcommand{\boxl}[1]{\hyperref[#1]{\textbf{Box~#1}}}

\newcommand{\avec}{\bm{\alpha}}
\newcommand{\dvec}{\bm{\psi}}

\newcommand{\mar}{\mathbf{T}}
\newcommand{\sta}{\bm{\pi}}

\newcommand{\ii}{\mathord{\text{\rmfamily\itshape i}}}
\newcommand{\pears}{\mathord{\text{\rmfamily\itshape r}}}
\newcommand{\vis}[1]{\textcolor{iscolor}{\textbf{#1}}}

\usepackage{titlesec}

\titlespacing*{\paragraph}{0pt}{0.25ex}{0.5em}

\usepackage[font=small,labelfont=bf]{caption}
\newcommand{\Stwo}{\textsc{s}\scalebox{0.78}{2}}   %
\newcommand{\Eend}{\textsc{end}}                     %
\definecolor{rev}{RGB}{190,20,30}   %
\usepackage{adjustbox}
\usepackage{placeins}
\title{Transformers Converge to Invariant Algorithmic\,Cores}

\author{%
  Joshua S. Schiffman \\
  New York Genome Center \\
  New York, NY, USA \\
  \texttt{joshseth@gmail.com}}

\newcommand{\pcrit}{p_{\mathrm{crit}}}
\graphicspath{{figures/}{supp/}}
\begin{document}
\raggedbottom %

\maketitle
{\let\thefootnote\relax\footnotetext{Current affiliation: Immunai, New York, NY, USA. Code: \url{https://github.com/joshseth/cores}.}}

\begin{abstract}
Training selects for behavior, not circuitry:
many weight configurations can implement the same function.
Studying any single trained neural network thus risks describing accidents
of one training run rather than the computation itself.
This work shifts focus from what transformers happen to do
to what they must do by extracting \emph{algorithmic cores},
compact subspaces that are necessary and sufficient for a task
and that recur across independently trained models.
Here, Algorithmic Core Extraction (ACE)
is introduced to isolate these subspaces, causally validate them,
and recover the algorithms they implement
across settings ranging from synthetic tasks to large-scale pretrained models.
Markov-chain transformers embed three-dimensional cores
in nearly orthogonal subspaces yet recover identical transition spectra.
Modular-addition transformers form compact cyclic cores at grokking
that later inflate under continued regularization,
redundantly distributing the same computation across many functionally equivalent modes.
This functional redundancy
is found to accelerate the transition from memorization to generalization,
yielding an inverse scaling law for grokking time.
In six language models spanning two orders of magnitude in scale
(GPT-2 Small/Medium/Large, LLaMA-3.1, Gemma-2, and Qwen2.5),
subject--verb agreement is governed by a single, steerable axis
that aligns across architectures.
Flipping this axis inverts grammatical number throughout open-ended generation.
For indirect object identification, a multi-step task, the same six models route computation through compact multi-dimensional cores at two sites, linked by a linear operator that transfers across model families.
Together these results suggest that beneath the apparent complexity
of trained transformers lies a simpler, shared computational structure,
and that targeting invariants
rather than parameterizations
may offer a more tractable path to mechanistic understanding and control.
\end{abstract}

\section{Introduction}

A key obstacle  to mechanistic interpretability~\citep{elhage2021mathematical,sharkey2025open}
is underdetermination:
while training constrains model behavior -- how inputs are mapped to outputs --
it generally does not constrain how behavior is realized internally.
This poses a fundamental challenge for interpretability: if mechanisms do not generalize across realizations, which explanations are real?

Such \emph{functional equivalence} is routinely observed among independently trained artificial neural networks,
and has been investigated in
loss landscape geometry and model merging~\citep{draxler2018essentially, garipov2018loss,ainsworth2022git},
the nonidentifiability of mechanistic circuits~\citep{meloux2025everything},
representational similarity~\citep{kornblith2019similarity},
and in the Rashomon effect~\citep{breiman2001two}.
Yet, this phenomenon is not restricted to neural networks
and has been explored across scientific disciplines.
In biology, it appears as \emph{degeneracy}~\citep{edelman2001degeneracy} (e.g., in the genetic code),
and in evolution as \emph{system drift}~\citep{schiffman2022system},
where the wiring of a gene network changes but its function does not.
In control theory, different \emph{realizations}~\citep{kalman1962canonical,kalman1963mathematical} induce identical observable dynamics,
and in physics, \emph{gauge symmetry} indicates that many mathematical descriptions represent the same state.
A natural response is to shift focus from individual realizations to equivalence classes,
studying the invariants shared across them.
If mechanistic explanations of language models are to generalize across random seeds~\citep{gurnee2024universal},
checkpoints, and architectures, they should be tethered to stable,
implementation-invariant quantities rather than idiosyncratic details that vary across training runs.

To explore this perspective,
Algorithmic Core Extraction (ACE) is introduced
to isolate \emph{algorithmic cores}:
low-dimensional subspaces that are necessary and sufficient for a task and shared across independent realizations.
Applying ACE across four settings of escalating complexity demonstrates that functionally equivalent models can converge on compact,
invariant mechanisms. In single-layer transformers~\citep{vaswani2017attention},
ACE recovers ground-truth Markov chain dynamics. In modular addition,
it isolates the emergence of rotational dynamics at grokking.
Finally, in six pretrained language models (spanning GPT-2, LLaMA-3.1, Gemma-2, and Qwen2.5)~\citep{radford2019language,grattafiori2024llama3herdmodels,gemma_2024,qwen2025qwen25technicalreport},
ACE identifies a shared, one-dimensional core that causally steers subject--verb agreement
during open-ended generation, and a pair of cores that perform indirect object identification (IOI).

\textbf{This work contributes:} (1) a conceptual framework
for mechanistic interpretability that shifts focus from
realization-specific circuitry to invariants; (2) ACE, a
method for isolating compact subspaces that are causally
necessary and sufficient for task performance; (3) evidence
that these cores isolate interpretable mechanisms; (4) a
theory linking functional equivalence, regularization, and
grokking time; (5) a steerable one-dimensional
subject--verb agreement core shared by six distinct LLMs;
and (6) a controllable pair of cores orchestrating IOI, shared across the same LLMs.

\section{Methods}
\label{sec:ace}

The structure--function relationship is often many-to-one, but how many
different structures can implement the same function?
In linear system theory this can be answered with the \emph{Kalman decomposition}~\citep{kalman1962canonical,kalman1963mathematical,anderson1966equivalence,kalman1969topics},
which guarantees the existence of a \emph{minimal realization} -- a dynamical system
that can be empirically recovered via \emph{balanced truncation}~\citep{moore2003principal}.
\emph{Algorithmic Core Extraction} (ACE) operationalizes this principle for transformers by first extracting activation subspaces that are both highly active and relevant, then causally validating them with ablations, and finally fitting operators to identify the computations they perform (Appendix~\ref{app:cores}).
\paragraph{Extract.}
Fix a transformer layer with hidden dimension $D$.
For $N$ inputs, let $\mathbf{H} \in \mathbb{R}^{N \times D}$
denote the mean-centered activations, with rows $\mathbf{h}_i^\top$,
and let $f : \mathbb{R}^D \to \mathbb{R}^K$
map activations to $K$ task-relevant outputs.
Stack the Jacobians as
$\mathbf{J} \coloneq
[(\partial f / \partial \mathbf{h}_1)^\top \cdots (\partial f / \partial \mathbf{h}_N)^\top]^\top
\in \mathbb{R}^{NK \times D}$.
ACE finds directions that are jointly active and relevant
by computing the SVD of their interaction:\footnote{When $NK\!\gg\!D$, use SVD of $\mathbf{L}^\top\boldsymbol{\Gamma} \in \mathbb{R}^{D\!\times\!D}$ instead, where $\mathbf{L}\mathbf{L}^\top\!\!=\! \mathbf{H}^\top\mathbf{H} \!+\! \varepsilon\mathbf{I}$ and $\boldsymbol{\Gamma}\boldsymbol{\Gamma}^\top\!\!=\! \mathbf{J}^\top\mathbf{J}$.}
\begin{equation*}
    \mathbf{H}\mathbf{J}^{\top} \;=\; \mathbf{U}\mathbf{\Sigma}\mathbf{V}^\top.
\end{equation*}
The singular values quantify the joint activity and relevance of each direction
and provide a principled criterion for rank selection.
The \emph{algorithmic core} is obtained by mapping the leading $r$ columns of $\mathbf{U}$
back into activation space:
\begin{equation*}
    \mathcal{C} \,\coloneq\, \mathrm{span}\!\left(\mathbf{H}^\top \mathbf{U}_r\right),
\end{equation*}
and QR decomposition yields
an orthonormal basis $\mathbf{Q} \in \mathbb{R}^{D \times r}$
and core projector $\mathbf{P} \coloneq \mathbf{Q}\mathbf{Q}^\top$.

\paragraph{Validate.}
A core is \emph{sufficient} if the projection $\mathbf{P}\mathbf{h}$
preserves task performance, and \emph{necessary} if its complement
$\mathbf{h} - \mathbf{P}\mathbf{h}$ reduces it to near chance.

\paragraph{Identify.}
A core's computational structure is recovered by examining its coordinates
$\mathbf{z} = \mathbf{Q}^\top \mathbf{h}$ directly,
or by fitting an operator $\mathbf{A}$
(e.g., $\mathbf{z}_{t+1} \approx \mathbf{A}\mathbf{z}_t$ by least squares)
and inspecting its spectrum.

\section{Algorithmic core necessity and sufficiency}
The central goal of this manuscript is to determine
whether low-dimensional subspaces, or \emph{algorithmic cores}, within higher-dimensional trained transformers
exist that are functionally necessary and sufficient for task performance.
If so,
are such cores shared across independently trained transformers,
and do they admit simple mechanistic characterizations?

\begin{figure}[h]
  \centering
  \includegraphics[width=\linewidth]{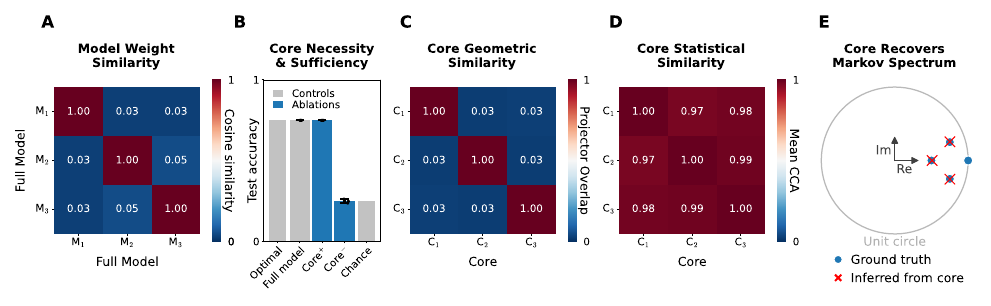}
  \caption{\textbf{Transformers trained on the same Markov task converge to a shared 3D causal core.}
  Three one-layer transformers were trained with different random seeds on next-token prediction for a four-state Markov chain (Appendix \ref{app:markov}).
  \textbf{(A)} Learned weights differ substantially across runs measured by cosine similarity.
  \textbf{(B)} 3D cores extracted from each 64D hidden state are necessary and sufficient under ablation (\tabl{tab:mark_ablations}),
  and compared to optimal $\sum_i \pi_i \max_j T_{ij}$ and chance $\max (\sta)$ theoretical controls, with transition matrix $\mar$ and stationary distribution $\sta$;
  points show individual accuracies and bars denote mean$\pm$sem.
  \textbf{(C)} Cores geometrically diverge with projector overlaps near zero and principal angles nearly orthogonal (\tabl{tab:mark_cca}).
  \textbf{(D)} Cores statistically align with mean cross-core CCAs reaching near unity (also see \tabl{tab:mark_cca}).
  \textbf{(E)} Dynamics fit in core recover the Markov chain nontrivial spectrum (\tabl{tab:mark_spec}).
  }
  \label{fig:markov}
\end{figure}

\paragraph{Recovering algorithmic cores.}
The analysis begins in a fully controlled setting:
three single-layer transformers ($d_{\rm model}=64$, $d_{\rm ff}=256$, $|V|=4$) trained with independent random seeds on a four-state Markov chain (Appendix~\ref{app:markov}).
Although each reached near Bayes-optimal test accuracy,
their learned weights exhibited near-zero cosine similarity,
indicating highly divergent parameterizations (\figpanel{fig:markov}{A}).
To search for a shared internal representation, ACE was applied to each model's 64-dimensional hidden state,
successfully isolating a 3-dimensional algorithmic core.
Ablations using all test data confirmed these cores were both necessary (removing the core drops accuracy to chance) and sufficient (retaining only the core preserves baseline accuracy) for the task (\figpanel{fig:markov}{B}; \tabl{tab:mark_ablations}; \hyperref[fig:sup_markov]{Figs.~\ref*{fig:sup_markov}} and~\ref{fig:sup_markov_energy}).

\paragraph{Geometric dissimilarity, statistical equivalence.}
To assess universality, each core recovered from the independently trained transformers was compared geometrically and statistically.
Despite meeting equivalent causal criteria, cores were embedded in nearly orthogonal subspaces:
projector overlap was $0.02$--$0.04$, and principal angles ranged from $75^\circ$--$90^\circ$ (\figpanel{fig:markov}{C}; \tabl{tab:mark_cca}).
Yet canonical correlation analysis (CCA)~\citep{morcos2018insights} revealed nearly exact statistical alignment,
with mean CCA correlations near $0.99$ (\figpanel{fig:markov}{D}; \tabl{tab:mark_cca}).
This suggests the cores encode the same information in different geometric coordinates -- a signature of functionally equivalent yet structurally divergent realizations.

\paragraph{Algorithmic cores encode Markov dynamics.}
To interpret what algorithm the cores implement,
a linear operator was fit to next-token dynamics
inside each core,
and relative to ``oracle'' prediction, these operators achieved strong fits: $R^2_{\rm core} / R^2_{\rm oracle} > 0.98$ (Appendix \ref{app:markov}).
Eigenvalues (the spectrum) of a linear operator determine its dynamics -- such as oscillations and growth rates -- so matching eigenvalues can indicate matching dynamics.
Remarkably, the eigenvalues of each fit operator
closely matched the non-trivial eigenvalues of the true Markov transition matrix (\figpanel{fig:markov}{E}; \tabl{tab:mark_spec}).
This suggests that the recovered cores learned to efficiently encode Markov dynamics:
trained transformers route inputs through a minimal, shared 3D subspace -- that is necessary and sufficient for performance -- and internally represent transition dynamics up to a change of coordinates.

\section{Algorithmic core emergence and evolution}
Because ACE is automated, it can recover learned computations without
presupposing their form and can trace how they evolve during training.
Modular addition is a natural test case: transformers trained on this task
exhibit grokking~\cite{power2022grokking,liu2022towards}, with high training
accuracy preceding a delayed spike in test accuracy. Prior work showed that
these models learn a Fourier ``clock'' algorithm~\cite{nanda2023progress},
but doing so required hypothesizing the mechanism a priori,  designing targeted probes,
and manually verifying circuits.

\begin{figure}[h]
  \centering
  \includegraphics[width=1.0\linewidth]{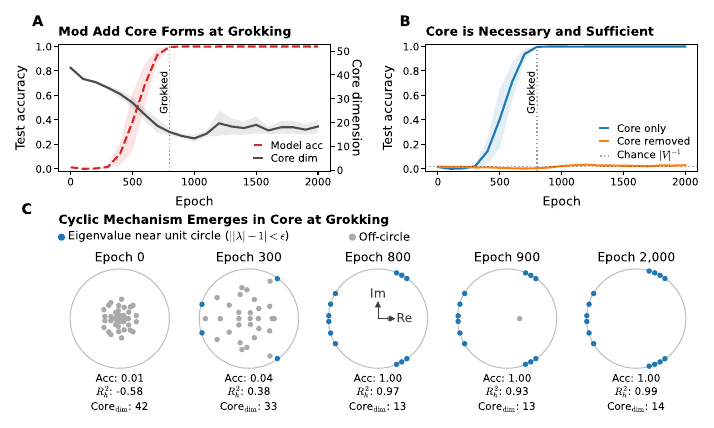}
  \caption{\textbf{Modular addition cores form at grokking and implement rotational mechanics.}
  Three two-layer transformers were trained with different random seeds.
  \textbf{(A)} Test accuracy exhibits grokking (red, mean$\pm$sem, left y-axis) coincident with algorithmic core formation (gray, mean$\pm$sem, right y-axis).
  \textbf{(B)} After grokking, the recovered cores are necessary and sufficient under ablation.
  \textbf{(C)} Automated operator fits in core coordinates reveal the emergence of a cyclic mechanism: before grokking, eigenvalues scatter inside the unit circle, while at grokking they snap onto it, indicating discovery of a rotational mechanism. $R_h^2$: held-out fit (Appendix~\ref{app:modadd}).}
  \label{fig:mod1}
\end{figure}
\paragraph{Cores crystallize at grokking.}
Three two-layer transformers
($d_{\rm model}=128$, $d_{\rm ff}=512$, $|V| = 53$)
were trained on modular addition ($a + b \equiv c \bmod 53$)
for $2 \times 10^3$ epochs under weight decay regularization to encourage generalization.
All models grokked: test accuracy remained near chance until spiking around epoch 800.
Coincident with this delayed generalization, algorithmic cores crystallized -- condensing into low-dimensional subspaces that are necessary and sufficient under ablation (\figpanel{fig:mod1}{A,B}; Appendix \ref{app:modadd}).

\begin{figure}[h]
  \centering
  \includegraphics[width=1.0\linewidth]{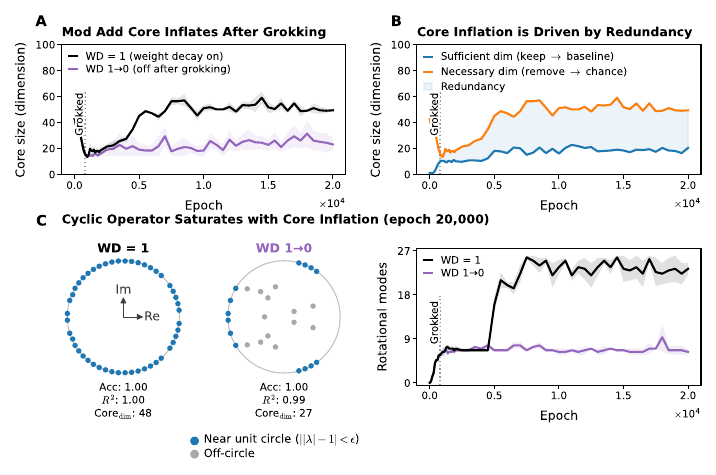}
  \caption{\textbf{Extended training with weight decay inflates cores.}
  Long-term training dynamics of transformers that grokked modular addition under different weight-decay schedules.
  \textbf{(A)} After grokking, core dimension continues to increase when weight decay is maintained (black, mean$\pm$sem), but remains compact when weight decay is disabled post-grokking (purple).
  \textbf{(B)} Core inflation is driven by redundancy: the number of dimensions sufficient to preserve performance is stable, while the number whose removal reduces performance to chance increases.
  Lines depict means across models trained with weight decay fixed.
  \textbf{(C)} (\textit{Left}) Dynamics fit in the terminal epoch reveal a saturated core operator when weight decay is maintained, in contrast to a more sparsely represented operator when weight decay is disabled.
  (\textit{Right}) Rotational modes (conjugate eigenvalue pairs) around the unit circle increase with extended training under weight decay, whereas when weight decay is removed, mode counts remain stable.
}
  \label{fig:mod2}
\end{figure}

\paragraph{Blind recovery of rotational dynamics in cores.}
At each checkpoint, a linear operator was fit to the second-layer ``shift'' (add 1) dynamics in each extracted core.
This revealed the emergence of a cyclic computational structure: at grokking, the operators' eigenvalues snap onto the unit circle (\figpanel{fig:mod1}{C}; \figpanel{fig:sup_modadd_seeds}{}), indicating rotational dynamics capable of modular addition.
Notably, this structure emerges directly from least-squares optimization in the core, without needing to prespecify an algorithmic form.
However, while all three models converged to cyclic operators, the specific rotational \emph{modes} (conjugate eigenvalue pairs) differed across runs -- another instance of functional equivalence without structural identity~\citep{chughtai2023toy,zhong2023clock,Olah2025ToyModelFaithfulness}.
Modular addition permits multiple valid modes and multiplicities, and models need not agree on which, nor how many, to use.
Remarkably, even at grokking, each operator contained more rotational modes than the single mode minimally required -- a hint of the redundancy that becomes extreme under extended training (\figpanel{fig:mod2}{}).

\paragraph{Cores inflate under extended training.}
Extending training to $2 \times 10^4$ epochs revealed an unexpected phenomenon: under continued weight decay, cores progressively inflated from approximately $15$ to $50$ dimensions.
In contrast, disabling weight decay post-grokking kept cores more compact (\figpanel{fig:mod2}{A}).
This inflation is driven by a pronounced increase in redundant encoding.
While the number of dimensions \emph{sufficient} for task performance remained stable, the number of dimensions \emph{necessary} to prevent chance-level performance expanded dramatically (\figpanel{fig:mod2}{B}; \figpanel{fig:sup_modadd_prefix}{}).
Operator analysis reveals how this transformer ``over-education'' manifests: under continued weight decay, operators accumulated rotational modes.
These approached the theoretical maximum of $\lfloor p/2 \rfloor = 26$ valid harmonic representations by the terminal epoch -- far exceeding the minimally required single mode (\figpanel{fig:mod2}{C}).
Disabling weight decay prevented this proliferation: cores remained compact, mode counts stayed sparse,
and operator structure remained stable. This
suggests that weight decay may actively drive the transition from parsimonious
algorithmic solutions to redundantly saturated representations.

\subsection{Redundancy drives core inflation and grokking}
\label{sec:theory}

That a regularization penalty designed to simplify representations
should instead inflate cores seems paradoxical.
The behavior, however, emerges naturally from minimizing the weight norm
within a highly redundant solution space.
Furthermore, this interplay between redundancy and regularization also predicts the timing of grokking itself.

\paragraph{Minimum norm requires maximum redundancy.}
After grokking, task loss is negligible and the gradient is dominated
by weight decay~\citep{varma2023explaining},
driving the network toward a minimum-norm solution.
By Fourier symmetry~\citep{chughtai2023toy},
modular addition mod $p$ admits $\lfloor p/2 \rfloor$
functionally equivalent modes,
each a 2D rotation with phase $\theta_k \coloneqq 2\pi k/p$.
Let $\avec \ge 0$ denote mode amplitudes
and $\dvec$ their label-contrasts,
where $\psi_k \coloneqq 1 - \cos\theta_k > 0$
is mode $k$'s contribution to the classification margin.
If modes are encoded in nearly orthogonal parameter
subspaces
then the weight norm satisfies $\|\mathbf{W}\|^2 \approx \|\avec\|^2$,
while correct classification requires margin
$\langle \avec, \dvec \rangle \ge \delta$.
Training thus implicitly solves
\begin{equation*}
    \min_{\avec \,\ge\, \mathbf{0}} \;\; \|\avec\|^2
    \qquad \text{subject to} \qquad
    \langle \avec, \dvec \rangle \ge \delta.
\end{equation*}
By the Cauchy--Schwarz inequality,
$\|\avec\|^2$ is minimized when $\avec \parallel \dvec$ --
that is, when \emph{every} mode is active --
with the optimal solution
$\avec^{\ast} = \left(\delta / \|\dvec\|_2^2\right)\dvec$.
Without orthogonality (e.g., in superposition~\citep{elhage2022superposition}), a mode overlap $\mathcal{S}\!\succ\!0$ (weight norm $\avec^\top\mathcal{S}\avec$) turns $\|\dvec\|_2^2$ into $\dvec^\top \mathcal{S}^{-1} \dvec$, reducing effective redundancy.
Weight decay thus acts as a redistribution force:
rather than simplifying the representation,
it spreads weight across all valid solutions.
Disabling weight decay removes this pressure,
consistent with observations in \figpanel{fig:mod2}{}.

\begin{figure}[h]
    \centering
    \includegraphics[width=0.74\linewidth]{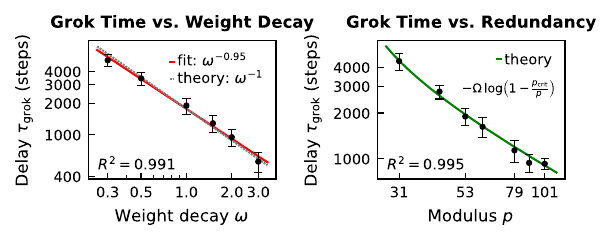}
    \caption{\textbf{Grokking time scales inversely with redundancy}{.} \textit{(Left)} Time to grok after memorization $\tau$ scales inversely with weight decay $\omega$, consistent with prior observations~\citep{liu2022omnigrok}: fit $\tau \propto \omega^{-0.95}$ ($R^2 = 0.991$), predicted $\omega^{-1}$. \textit{(Right)} $\tau$ against redundancy $p$: ODE fit $R^2 = 0.995$ ($\hat\Omega \approx 3{,}800$, $\hat p_{\mathrm{crit}} \approx 21$); $p = 17$, 13, and 11 failed to grok after 50{,}000 steps. Points: mean$\pm$sd over 12 seeds (Appendix~\ref{app:grok_sweep}).}
    \label{fig:grok_sweep}
\end{figure}

\paragraph{Functional equivalence accelerates grokking.}
The same redistribution pressure governs the speed of grokking.
Define the grokking delay
$\tau_{\mathrm{grok}} \coloneqq \tau_{\mathrm{gen}} - \tau_{\mathrm{mem}}$
as the time between memorization and generalization,
and model the transition to generalization as the margin
$m(t) \coloneqq \langle \avec(t), \dvec \rangle$ reaching threshold $\delta$.
After memorization, with task gradients largely vanished
and weight decay ($\omega$) dominating,
the expected margin trajectory follows (Appendix \ref{app:grok_math})
\begin{equation*}
    \dot{m}(t) \;=\; -\omega\,m(t) \;+\; c\,\omega\,\|\dvec\|_2^2.
\end{equation*}
Crucially, functional equivalence makes the margin-driving direction
additive across modes, giving\footnote{%
For odd $p$, using $\sum_{k=1}^{\lfloor p/2 \rfloor}\cos\theta_k
= \sum_{k=1}^{\lfloor p/2 \rfloor}\cos 2\theta_k = -\tfrac{1}{2}$
and expanding $(1-\cos\theta_k)^2$ gives
$\|\boldsymbol{\psi}\|_2^2 = \tfrac{3}{4}p$.}
$\|\boldsymbol{\psi}\|_2^2 \propto p$.
Each redundant mode amplifies the mean-drift velocity
toward generalization,
consistent with the multiple active modes observed at grokking
(\figpanel{fig:mod1}{C}).
When $p < d_{\mathrm{model}}$
the initial memorized solution has negligible margin
($m(0) \approx 0$),
and grokking occurs when $m(\tau) = \delta$.
Solving for the expected grokking time delay yields an expression
that linearizes for high redundancy ($p \gg p_{\mathrm{crit}}$)
into a simple inverse scaling law:
\begin{equation*}
    \tau_{\mathrm{grok}}(p)
    \;=\; -\,\Omega\,\log\!\left(1 - \frac{p_{\mathrm{crit}}}{p}\right)
    \;\approx\; \frac{\Omega\, p_{\mathrm{crit}}}{p}
    \;\propto\; \frac{1}{\omega\, p}.
\end{equation*}
Two empirical constants govern this expression:
an \emph{optimizer constant} $\Omega \propto \omega^{-1}$
that sets the timescale of grokking when it occurs,
and an \emph{architectural constant} $p_{\mathrm{crit}}$
that determines whether it can occur at all.
Grokking time thus shrinks with both weight decay and functional redundancy.
These predictions are validated by sweeping $\omega$ and $p$ in transformers
(\figpanel{fig:grok_sweep}{}; \hyperref[fig:grok_one_layer]{Figs.~\ref*{fig:grok_one_layer}} and~\ref{fig:grok_controls}; Appendix \ref{app:grok_sweep}).

\paragraph{Summary.}
The algorithmic core framework -- automated operator extraction from causally defined,
low-dimensional core subspaces -- can mechanistically characterize and trace the evolution of computations transformers learn throughout training.
In modular addition, the extracted cores exhibit rotational dynamics consistent with the task's cyclic structure,
crystallize at grokking, and inflate under extended weight decay.
This inflation reflects transformers converging on the optimal weighting strategy under regularization:
to distribute weight across all functionally equivalent representations.
This same pressure -- regularization utilizing redundancy -- predicts the speed of grokking,
explaining the transition from memorization to generalization.
The next question is whether these tools scale to larger and more complex systems.

\begin{figure}[h]
  \centering
  \includegraphics[width=\linewidth]{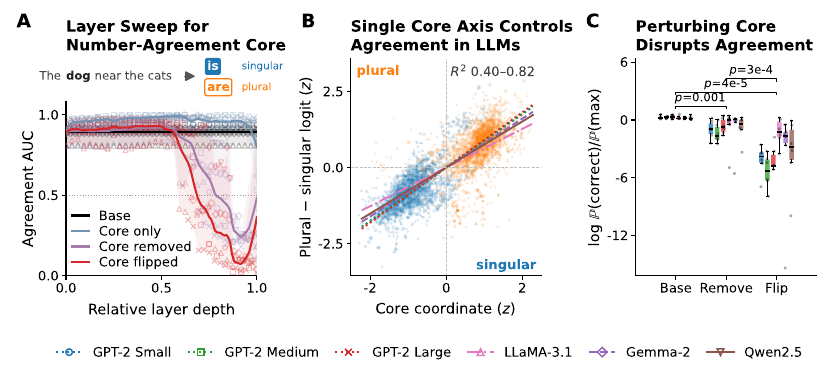}
  \caption{\textbf{Subject--verb agreement is mediated by a shared 1D core across LLMs.}
  The core framework was applied to GPT-2 Small, Medium, Large, Gemma-2, LLaMA-3.1, and Qwen2.5 to isolate a low-dimensional mechanism for number agreement.
  \textbf{(A)} Layer sweep: agreement performance (AUC) vs.\ normalized layer depth, averaged across LLMs (lines) with per-model measurements (markers) and shaded min--max bands.
	Agreement performance is the probability that the model assigns a higher plural-vs-singular verb-preference score to a plural than to a singular prompt.
	\textbf{(B)} Projecting last-token hidden states onto the core produces a nearly linear control axis for the singular--plural logit margin; per-model affine fits are shown after $z$ scoring.
  \textbf{(C)} Removing the core degrades agreement, while flipping it inverts verb preference.
  Box plots summarize prompt-level agreement scores under perturbation; reported $p$-values combine per-model paired Wilcoxon tests using Fisher's method.}
  \label{fig:gpt2}
\end{figure}
\begin{figure}[h]
\centering
\footnotesize
\newcommand{\gp}[1]{\textcolor{black!50}{#1}}
\noindent\rule{\linewidth}{0.4pt}\\[1pt]
\noindent\textbf{GPT-2 Medium}\hfill Prompt: \emph{We hold these truths to be self-evident:}\\[2pt]
\begin{minipage}[t]{0.485\linewidth}\raggedright
\textbf{Base}\\
\gp{We hold these truths to be self-evident:} that all men \textbf{are} created equal, that they \textbf{are} endowed by their Creator with certain unalienable Rights [i.e., without a priori moral rights], and that among these \textbf{are} Life, Liberty and the pursuit of Happiness.''
\end{minipage}\hfill
\begin{minipage}[t]{0.485\linewidth}\raggedright
\textbf{Core Steering}\\
\gp{We hold these truths to be self-evident:} that all men \vis{is} created equal, that they \vis{is} endowed by their Creator with certain unalienable Rights [i.e., the right to life], that among these \vis{is} Life[.]''
\end{minipage}\\[3pt]
\noindent\rule{\linewidth}{0.4pt}\\[1pt]
\noindent\textbf{LLaMA-3.1-8B}\hfill Prompt: \emph{As a new field of research, artificial intelligence}\\[2pt]
\begin{minipage}[t]{0.485\linewidth}\raggedright
\textbf{Base}\\
\gp{As a new field of research, artificial intelligence} \textbf{has} already delivered us with its revolutionary ways to solve complex medical issues. Its potential to address more such problems can boost the healthcare system...
\end{minipage}\hfill
\begin{minipage}[t]{0.485\linewidth}\raggedright
\textbf{Core Steering}\\
\gp{As a new field of research, artificial intelligence} \vis{have} already made great strides to improve our lives. AI has been instrumental in providing a more efficient... But how will we know what we \vis{has} the potential to do...
\end{minipage}\\[3pt]
\noindent\rule{\linewidth}{0.4pt}

\caption{\textbf{Steering the core induces systematic agreement violations in open-ended text generation.}
Each panel shows text generated by select language models from the same prompt under two conditions: \emph{Base} and \emph{Core Steering} (activations adaptively modulated).
Colored words highlight selected agreement violations.
}
\label{fig:sva_gen}
\end{figure}
\section{Scaling ACE to LLMs: A universal 1D core}
\label{sec:gpt2}
To test whether ACE scales beyond synthetic tasks to production-scale models,
cores were extracted from six pretrained language models spanning four families and two orders of magnitude in scale --
GPT-2 Small, Medium, and Large (124M--774M parameters)~\citep{radford2019language,wolf2020transformers},
LLaMA-3.1 (8B)~\citep{grattafiori2024llama3herdmodels}, Gemma-2 (9B)~\citep{gemma_2024}, and Qwen2.5 (32B)~\citep{qwen2025qwen25technicalreport} --
which differ in architecture, training corpus, and tokenization.
The target task is \emph{subject--verb number agreement},
a tractable linguistic computation with clear ground-truth labels (singular vs.\ plural subject) and a well-defined behavioral output (verb selection);
admitting systematic evaluation via controlled prompts (\tabl{tab:sup_sva_classes}) and a scalar verb-preference score~\citep{linzen2016assessing,marvin2018targeted,finlayson2021causal} (Appendix \ref{app:sva}): the final-token logit margin $f = (\ell_{\textit{are}} + \ell_{\textit{were}}) - (\ell_{\textit{is}} + \ell_{\textit{was}})$.

\paragraph{Localizing a shared 1D agreement core.}
To localize the agreement mechanism,
candidate cores were extracted at each layer and evaluated via causal ablations.
Across all six models, early layers exhibited minimal causal influence,
but a highly potent core consistently emerged in the late layers (\figpanel{fig:gpt2}{A}).
At the layer of maximal effect, this agreement core is remarkably one-dimensional -- a single axis separated from all remaining directions by a large spectral gap (\tabl{tab:gpt2_sizes}), even though a scalar margin across nonlinear layers neither bounds the Jacobian rank nor guarantees that a 1D subspace will be causally necessary (cf.\ Section~\ref{sec:ioi_main}; Appendix~\ref{app:sva_controls}).

\paragraph{Causal validation and control.}
Observationally, this axis behaves as a graded number coordinate: projection onto it predicts the singular--plural logit margin across models (\figpanel{fig:gpt2}{B}),
aligning with the linear representation hypothesis~\citep{park2023linear} and with low-rank accounts of network computation~\citep{mastrogiuseppe2018linking}.
However, because subspace projections alone can be deceptive~\citep{belinkov2022probing,makelov2023subspace}, claims here are strictly grounded in causal ablations.
Despite its compact size, this single axis is
sufficient (retaining it preserves agreement; AUC $\ge 0.91$),
necessary (removing it collapses agreement below chance; AUC $\le 0.25$; Appendix~\ref{app:sva}; \figpanel{fig:sup_jacobian}{}),
and directionally controllable.
Reflecting activations through this axis inverts verb preferences, inducing strong disagreement with the subject (AUC $\le 0.10$; \tabl{tab:gpt2_ablations}).
At the prompt level, for instance, core inversion on ``The key next to the cabinets'' drives $\mathbb{P}(\textit{is})$ from $0.51$ down to $0.01$, while boosting $\mathbb{P}(\textit{are})$ from $0.06$ to $0.75$.
Compared with the unembedding direction and PCA, only ACE is both necessary and sufficient;
however, ACE cores largely overlap with contrastive-steering directions, revealing that contrastive steering is a special case of rank-1 ACE (\hyperref[fig:sup_sva_controls]{Figs.~\ref*{fig:sup_sva_controls}} and~\ref{fig:sup_sva_keep_remove}; Appendix \ref{app:cores}).

\paragraph{Alignment across LLMs.}
Projecting last-token hidden states onto each model's agreement
core (\figpanel{fig:gpt2}{B})
yields a signed grammatical-number coordinate that tracks verb preference.
Because cores are one-dimensional,
cross-model alignment reduces to fixing a sign convention
and comparing projected coordinates.
Within the GPT-2 family -- three models that share an architecture
and training procedure -- coordinates align tightly
(Pearson's $\pears = 0.92$--$0.97$).
More strikingly, alignment persists across families:
between Qwen2.5, LLaMA-3.1, Gemma-2, and the GPT-2 models,
correlations range from $0.62$ to $0.94$
and remain positive after regressing out subject and attractor number
(\hyperref[fig:sva_cca]{Figs.~\ref*{fig:sva_cca}} and~\ref{fig:sup_nulls}; Appendix~\ref{app:sva}).
The strongest cross-family correlations
approach the within-family ceiling,
indicating that the agreement core encodes grammatical number
in a way that is largely independent of architecture,
tokenization, training corpus, and scale.

\paragraph{Adaptive core steering inverts grammar in open-ended generation.}
A stronger test of the core's role is whether it governs agreement
throughout autoregressive generation,
where each token conditions subsequent predictions.
To test this, the core-axis intervention was applied \emph{adaptively}
at each decoding step.
Modulating interventions by token sensitivity
induced systematic agreement violations in all six models (\figpanel{fig:sva_gen}{}; Appendix~\ref{app:sva}).
Singular subjects recruited plural verbs,
plural contexts shifted toward the singular,
and errors cascaded as toggling the number variable corrupted downstream predictions.
Crucially, the effect generalized well beyond the specific verbs (\textit{is/are/was/were})
used to define the initial preference score.
The emergence of agreement failures in entirely different word classes supports the
interpretation that the core encodes a global grammatical-number variable, rather than a narrow, verb-specific heuristic.

\section{Multi-step computation flows through connected cores}
\label{sec:ioi_main}
Many linguistic behaviors may require composing multiple interconnected steps.
Indirect object identification (IOI) is a canonical benchmark for such computation:
in ``When Mary and John went to the store, John gave a drink to'', predicting \emph{Mary} requires tracking names,
detecting repetition, and routing.
Circuit tracing identified attention-head circuits implementing this algorithm in GPT-2 Small (26 heads) and Medium~\citep{wang2022interpretability,merullo2024circuit},
but such methods have not generalized to larger models~\citep{haklay2025position,franco2026prompt}.
To test whether ACE can isolate a compact, cross-architectural mechanism for a multi-step task,
candidate cores were extracted across layers in the same six LLMs,
using role-swap prompt pairs that differ only in which name repeats (Appendix~\ref{app:ioi}).

\paragraph{A single-site core reveals distributed computation.}
Extracting a core at the last ({\Eend}) token at a consistent relative depth (0.61--0.88)
in all six LLMs
isolates a 7--21-dimensional subspace
that is sufficient and controllable, but \emph{not necessary}.
Reflecting activations through the core inverts the predicted name (AUC $\approx 0$),
and transplanting the counterfactual state steers the model to the opposite answer ($\approx 0$).
However, removal leaves the behavior largely intact
(\tabl{tab:review_single}; \hyperref[fig:ioi_single_core]{Figs.~\ref*{fig:ioi_single_core}} and~\ref{fig:sup_ioi_single}),
suggesting this multi-step computation may be mediated by multiple cores distributed across layers.

\begin{figure}[t]
\centering
\includegraphics[width=\linewidth]{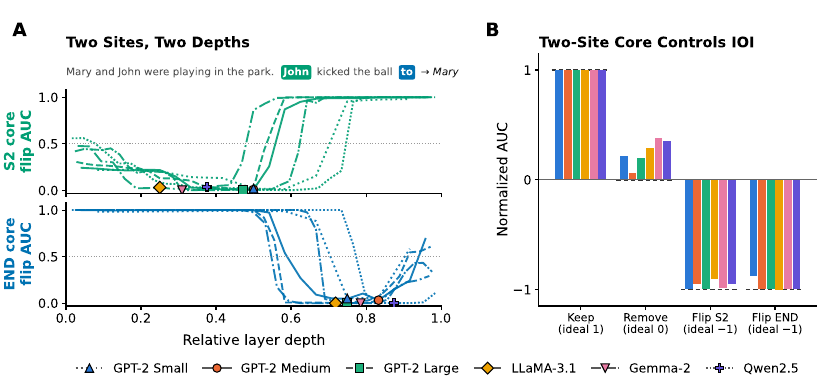}
\caption{\textbf{IOI is carried by cores at two sites in six models.} (A) Layer sweep on the extraction set: flip AUC (low =
the answer flips) of each site's core flipped alone, at rank $\max(16,\ 2.5\%\text{ of }D)$, against relative depth. The repeated
name ({\Stwo}) and the final token ({\Eend}) each have one band per model; markers: chosen layers. (B) The smallest core per model
(16--143 dimensions in total; \tabl{tab:ioi_cores}) on test prompts (held-out frames, 40 new names): keep and remove at both sites and
flip at each site alone, as normalized AUC, $(\mathrm{AUC}-0.5)/(\mathrm{AUC}_{\mathrm{clean}}-0.5)$ (ideal 1, 0, $-1$).}
\label{fig:ioi_layers}
\end{figure}

\paragraph{Two-site cores localize and control the multi-step mechanism.}
To capture this more distributed computation,
cores were extracted at two sites: the repeated name ({\Stwo}) and final ({\Eend}) tokens,
using within-pair differences across role-swap prompts to isolate task-driven activation variance from background prompt variance,
and localized to layer depths 0.25--0.50 and 0.72--0.88 (\figpanel{fig:ioi_layers}{A}; Appendix~\ref{app:ioi}).
Across all six models, the sufficient two-site core --
at each site the smallest rank that alone preserves the behavior and inverts it when flipped -- totals just 6--51 dimensions.
The smallest core pair that is also necessary (removal AUC near chance) totals 16--143 dimensions (1.6--4.8\% of $D$; \tabl{tab:ioi_cores}; \figpanel{fig:ioi_curves}{}).
On held-out test prompts with unseen sentence frames and names,
retaining this two-site core preserves the behavior (AUC $> 0.99$),
flipping either site alone inverts the predicted name (AUC $< 0.07$),
and removing both sites degrades performance toward chance
(0.53--0.69, \figpanel{fig:ioi_layers}{B}; 0.52--0.71 on the prompts of Wang et al.~\citep{wang2022interpretability}, \tabl{tab:ioi_heldout}),
in contrast to controls (\tabl{tab:ioi_controls}; \figpanel{fig:ioi_wang_controls}{}).
The dimensions required to fully disrupt the task outnumber those needed to perform it,
perhaps a consequence of core inflation, as was observed in modular addition (\figpanel{fig:mod2}{B}).

\paragraph{A reduced linear model links cores across architectures.}
To mechanistically characterize how these two cores compose,
least-squares operators were fit to core activations on the extraction set:
a transition matrix $\mathbf{A}$ maps the centered {\Stwo} core state to the {\Eend} core state,
and the {\Eend} core is projected onto the name-preference logit margin $m$,
\begin{equation}
\hat{\mathbf{z}}_E = \bar{\mathbf{z}}_E + \mathbf{A}\,(\mathbf{z}_S - \bar{\mathbf{z}}_S), \qquad \hat m = \mathbf{c}^\top \hat{\mathbf{z}}_E + d,
\label{eq:ioi_main}
\end{equation}
where $\mathbf{z}$ are core coordinates and $\bar{\mathbf{z}}$ their fixed means on the extraction set.
Fit without any intervention, and with the {\Stwo} core state as its only input,
this two-stage model predicts the right name on held-out sentences and new names (AUC $> 0.99$).
At {\Stwo}, a single axis -- the mean within-pair difference -- suffices to decode the answer but not to control it reliably:
adding this axis plus the difference between the two names to the {\Stwo} core, without partner activations,
reverses the answer in all six models (AUC $< 0.03$), whereas neither part alone reliably does so (0.45--1 and 0.31--0.98).
Moreover, without having seen a swap, $\mathbf{A}$ predicts 61--87\% of the change in the {\Eend} core when the {\Stwo} core is swapped
(about 0\% when fit on shuffled pairs), removing 55--78\% of the error left by the best single fixed direction.
Remarkably, this inter-site dynamics operator is shared across model families: transferring
$\mathbf{A}$  from one language model to another
via natural activation alignments
predicts 60\% of the target's {\Eend} core response to an {\Stwo} swap on average,
approaching the 77\% explained by the target's own $\mathbf{A}$ (0\% with shuffled alignment; Appendix~\ref{app:ioi}, \tabl{tab:ioi_reduced}).

\section{Discussion}
These results suggest that transformer computations may be governed
by low-dimensional mechanisms that recur across independent training runs
despite substantial variation in learned parameters.
These findings have implications for how we conceptualize mechanistic interpretability.

\paragraph{Invariance, not sparsity or circuitry.} Mechanistic interpretability has largely studied implementations, such as circuits of attention heads and neurons~\citep{elhage2021mathematical,olah2020zoom,wang2022interpretability,ameisen2025circuit,lindsey2025biology},
or sparse decompositions of activations into interpretable features~\citep{cunningham2023sparse,bricken2023monosemanticity,dunefsky2024transcoders,templeton2024scaling}.
Such descriptions can be highly precise, but they face a conceptual challenge:
they may be implementation-specific.
Two models might compute the same function using entirely different circuits and coordinate systems~\citep{meloux2025everything,fel2025archetypal}.
The core framework shifts the explanatory target from implementation to invariant.
The motivation for sparse features parallels a classical aim in linear algebra: diagonalization.
But the fundamental power of diagonalization lies not in sparsity per se,
but in revealing invariants --
eigenvalues preserved under change of basis.
Sparsity is basis-dependent; invariants are not.
Likewise, circuits and sparse features describe coordinates of implementation,
while cores identify the causal subspaces and dynamics preserved across implementations.
Where features or circuits recur across models,
perhaps identified via cross-coders~\cite{lindsey2024sparse},
the approaches converge.
Where they diverge, invariance provides a reliable criterion for distinguishing structural essence from artifact.

\paragraph{Cores as internal world models.}
The observation that independent models converge to the same invariant structure raises a natural question:
what anchors these shared representations? If these cores are not artifacts of the architecture or training run,
they might reflect the data-generating process itself.
When algorithmic cores recover ground-truth task structure --
Markov transition spectra, cyclic operators for modular arithmetic --
they encode not merely input--output mappings but internal representations of
the generative process underlying the data~\citep{li2022emergent,gurnee2023language,huh2024platonic}.
This aligns with two classical ideas:
the \emph{good regulator theorem}~\citep{conant1970every}
and the \emph{internal model principle}~\citep{francis1976internal}
from control theory, which hold that a system that regulates its environment well
must contain a model of it.
When a core is isomorphic to the task-generating process,
interpretability may be viewed as a form of internal-model recovery.
\paragraph{Redundancy accelerates grokking.}
Post-memorization, weight decay drives parameters along the degenerate zero-loss manifold~\citep{bushnaq2024using} toward a minimum-norm solution,
accumulating margin drift across functionally equivalent modes to accelerate generalization.
Crossing the discrete classification threshold produces a sharp jump in test accuracy even though the weight trajectory stays smooth; speculatively, scaling may likewise produce capability jumps once models can realize large classes of functionally equivalent solutions.
An analogous phenomenon appears in evolutionary genetics, where robustness creates extended networks of phenotype-preserving genotypes that facilitate the discovery of new functions via neutral drift~\citep{wagner2008robustness,wagner2012role}.
A practical consequence is that cores are most compact immediately after grokking and subsequently inflate.
This suggests a natural \emph{interpretability window}: annealing weight decay toward zero shortly after task convergence may help preserve the most compact solution.

\paragraph{System drift and model merging.}
System drift describes how a gene network can preserve its phenotype
while its underlying genetic wiring diverges, effectively drifting through a neutral space.
Because the set of functionally equivalent realizations is not generally convex or closed under recombination,
mixing divergent solutions often produces \emph{hybrid incompatibility}~\citep{schiffman2022system}.
Transformers exhibit an analogous pattern:
models trained from different initializations implement equivalent cores
embedded in nearly orthogonal subspaces,
revealing substantial representational drift despite functional equivalence.
This orthogonality implies that na\"ive weight interpolation
between geometrically divergent models moves off the solution manifold,
consistent with empirical difficulties in model merging~\citep{garipov2018loss,ainsworth2022git}.
By contrast, extracting and aligning algorithmic cores may offer a principled diagnostic for merge-compatibility
and a potential coordinate system for successful recombination.

\paragraph{Limitations and future directions.}
Extracting cores from multifunctional models is hypothesis-driven
and requires framing a precise and falsifiable mechanistic question;
systematic approaches to scale this remain open.
More fundamentally, necessity, sufficiency, and control of a subspace are an incomplete vocabulary for describing mechanism in a complex system.
The extraction procedure itself admits natural extensions:
nonlinear dimensionality reduction in place of the active component,
learned probes in place of Jacobians,
and Koopman operator approximations~\citep{brunton2021modern}
for tasks with nonlinear dynamics.
More broadly, the relevant invariants for complex tasks
are not obvious \emph{a priori};
future work might discover them empirically
by asking what core properties are shared across independently trained models.
Finally, methods that identify causally effective subspaces may enable more targeted model control:
this could support auditing and debugging, but also creates misuse risks if used to induce systematic errors or circumvent intended behaviors.

\paragraph{Conclusion.}
The results in this work point toward a view of transformer computation as organized around low-dimensional invariants:
subspaces that are preserved across training runs, necessary and sufficient for task performance,
and structured in ways that mirror the tasks themselves.
If this view is approximately correct, interpretability efforts may benefit from targeting such invariants --
seeking the computational essence that recurs across realizations rather than cataloging the implementation details that drift across seeds and architectures.
Algorithmic Core Extraction is a first operationalization of this principle; extending it from multi-step linguistic routing to open-ended reasoning and safety-critical behaviors is a central frontier for future work.
The guiding principle -- focus on what is preserved, not what is particular -- may prove durable.

\section*{Acknowledgments and disclosures}
I would like to thank Drs.~Alison Pickover and Dan Landau for their support. This work received no external funding. The author declares no competing interests.

\section*{Code availability}
Code to reproduce all analyses and figures is available at \url{https://github.com/joshseth/cores}.

\bibliographystyle{unsrt}
\bibliography{refs}

\section*{Appendix}
\appendix
\label{app:app}
\renewcommand{\thetable}{A\arabic{table}}
\setcounter{table}{0}
\renewcommand{\thefigure}{A\arabic{figure}}
\setcounter{figure}{0}

\section{Algorithmic core extraction}
\label{app:cores}

\paragraph{Functional equivalence and minimal realizations.}
The structure--function relationship is often many-to-one~\citep{bellman1970structural}:
there is more than one way to realize a behavior.
But how many different structures can realize identical input--output functions?
Can the space of functionally equivalent structures be characterized?

In linear system theory, this question has an exact answer~\citep{kalman1962canonical}.
Consider a linear time-invariant system with hidden state $\mathbf{x} \in \mathbb{R}^n$,
input $\mathbf{u} \in \mathbb{R}^m$, and output $\mathbf{y} \in \mathbb{R}^\ell$:
\begin{align*}
    \dot{\mathbf{x}} &= \mathbf{A}\mathbf{x} + \mathbf{B}\mathbf{u}, \\
    \mathbf{y} &= \mathbf{C}\mathbf{x}.
\end{align*}
The system's input--output behavior is fully determined by its \emph{impulse response}
$\boldsymbol{\zeta}(t) \coloneq \mathbf{C}e^{\mathrm{\textbf{A}}t}\mathbf{B}$.
Two systems with different weights $(\mathbf{A}, \mathbf{B}, \mathbf{C})$ and
$(\tilde{\mathbf{A}}, \tilde{\mathbf{B}}, \tilde{\mathbf{C}})$
are functionally equivalent if they produce identical outputs for all inputs -- that is,
if their impulse responses match ($\boldsymbol{\zeta}(t) = \tilde{\boldsymbol{\zeta}}(t)$).

For systems of equal dimension, functional equivalence corresponds exactly to coordinate change:
$(\mathbf{A}, \mathbf{B}, \mathbf{C})$ and
$(\mathbf{V}\mathbf{A}\mathbf{V}^{-1}, \mathbf{V}\mathbf{B}, \mathbf{C}\mathbf{V}^{-1})$
share the same impulse response for any invertible $\mathbf{V}$.
But systems of different sizes can also be functionally equivalent
if some internal states are either unreachable (unaffected by input)
or unobservable (irrelevant to the output).

The \emph{Kalman decomposition} makes this precise,
partitioning any system's state space into four subspaces
according to reachability and observability~\citep{kalman1962canonical, kalman1963mathematical,anderson1966equivalence,kalman1969topics}.
Only states that are both reachable and observable contribute to input--output behavior;
the rest represent degrees of freedom that can vary without affecting function.
This decomposition guarantees the existence of a \emph{minimal realization} --
the smallest-dimensional system that reproduces an input--output map,
unique up to coordinate change -- and enables extracting it.
These results from system theory conceptually motivate the methods
developed in this manuscript.

\paragraph{Algorithmic core extraction.}
The goal here is an analogous decomposition for transformers.
The Kalman decomposition provides an exact algebraic characterization for linear systems;
for transformers, no such closed-form decomposition exists,
but the principle can be applied empirically:
identify directions that are both input-driven (active) and output-relevant (relevant).
If the system were linear, this would reduce to \emph{balanced truncation}~\citep{moore2003principal},
a technique in model reduction that finds coordinates
in which reachability and observability are aligned.

Here, \textbf{ACE} (\emph{Algorithmic Core Extraction})
operationalizes this approach for artificial neural networks.
Let $\mathbf{H}\in\mathbb{R}^{N\times D}$ denote mean-centered hidden activations
at a transformer layer of interest,
with rows $\mathbf{h}_i^\top\in\mathbb{R}^{1\times D}$ for each of $N$ inputs,
to define \emph{active} directions.
To quantify \emph{relevant} directions,
let $f\colon \mathbb{R}^D \to \mathbb{R}^K$ map activations to task-relevant outputs
and let $\mathbf{J}\in\mathbb{R}^{NK\times D}$ stack the $N$ Jacobians
$\partial f/\partial \mathbf{h}_i \in \mathbb{R}^{K\times D}$ as row blocks.

To find directions that are jointly active and relevant,
ACE computes the SVD of their interaction:
\begin{equation*}
    \mathbf{H}\mathbf{J}^{\top}
    \;=\;
    \mathbf{U}\mathbf{\Sigma}\mathbf{V}^\top.
\end{equation*}
The singular values quantify the joint importance of each direction,
providing a principled criterion for rank selection.
Let $\mathbf{U}_r$ denote the first $r$ columns of $\mathbf{U}$.
The \emph{algorithmic core} is the subspace obtained by projecting
these interaction modes back into activation space,
\begin{equation*}
    \mathcal{C} \,\coloneq\, \mathsf{\mathrm{span}}\!\left(\mathbf{H}{^\top}\mathbf{U}_r\right).
\end{equation*}
The core's orthonormal basis $\mathbf{Q}\in\mathbb{R}^{D\times r}$ is given by the QR decomposition
\begin{equation*}
    \mathbf{H}^\top \mathbf{U}_r \;=\; \mathbf{Q}\mathbf{R},
\end{equation*}
and thus, the \emph{core projector} is defined as $\mathbf{P} \coloneq \mathbf{Q}\mathbf{Q}^\top$.

\textit{Note on implementation.}
Computing $\mathbf{H}\mathbf{J}^\top\in\mathbb{R}^{N\times NK}$ is unnecessary (and inefficient when $NK\gg D$).
Instead form the activation covariance $\mathbf{A}\coloneq \mathbf{H}^\top \mathbf{H}\in\mathbb{R}^{D\times D}$
and sensitivity matrix $\mathbf{S}\coloneq \mathbf{J}^\top \mathbf{J}\in\mathbb{R}^{D\times D}$.
Take square-root factors
$\mathbf{A}+\varepsilon\mathbf{I}=\mathbf{L}\mathbf{L}^\top$ and $\mathbf{S}=\mathbf{\Gamma}\mathbf{\Gamma}^\top$,
then compute the SVD of the resulting $D\times D$ matrix
$\mathbf{L}^\top \mathbf{\Gamma} \;=\; \mathbf{U}\mathbf{\Sigma}\mathbf{V}^\top,$
which yields the core subspace:
$\mathsf{\mathrm{span}}(\mathbf{L}\mathbf{U}_r).$

\paragraph{Causal validation.}
The core is validated through ablation,
with $\tilde{\mathbf{h}}$ denoting the activation after intervention:
\begin{align*}
    \text{Core-only (to test sufficiency):} \quad & \tilde{\mathbf{h}} = \mathbf{P}\mathbf{h}, \\
    \text{Core-removed (to test necessity):} \quad & \tilde{\mathbf{h}} = \mathbf{h} - \mathbf{P}\mathbf{h}.
\end{align*}
A subspace is deemed \emph{sufficient} if core-only preserves task performance,
and \emph{necessary} if core-removed reduces performance to approximately chance.
The energy-based rank can be refined by finding the minimal $r$
such that keeping only the core maintains baseline accuracy and removing it drops accuracy to near chance.
Both tests are joint properties of the subspace under this projection: they do not imply that each core dimension is individually necessary, that the orthogonal complement carries no task information, or that the intervention stays in distribution.

\begin{figure}[H]
\centering
\includegraphics[width=\linewidth]{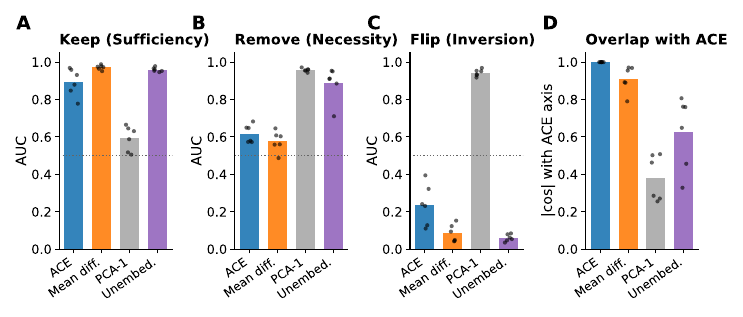}
\caption{\textbf{Subject--verb agreement: causal control and axis geometry} across six models (bars: mean; points: models). Keep, remove, and flip
AUC for ACE, the difference of means, PCA-1, and the unembedding direction, and their $|\cos|$ with the ACE axis. Prompts: subject and attractor number vary independently.}
\label{fig:sup_sva_controls}
\end{figure}

\paragraph{Ablations test ACE; they do not construct its basis.}
ACE orders directions in closed form from activations and Jacobians alone (the SVD above), without observing an ablation.
In the Markov task the rank was fixed by the 99.9\% energy criterion (Appendix~\ref{app:markov}), and in subject--verb agreement by the spectral gap (Appendix~\ref{app:sva}), so the ablations there test an energy-selected subspace.
The same tests fail where the structure is absent or incomplete:
rank-matched random subspaces fail them in the Markov task (\figpanel{fig:sup_markov}{});
the unembedding direction is sufficient but not necessary for agreement (\figpanel{fig:sup_sva_controls}{});
before grokking the modular-addition core is neither (\figpanel{fig:mod1}{B});
and in IOI a single-site core is sufficient but, at the size set by flipping, not necessary (core-removed AUC 0.80--1; \tabl{tab:review_single}).
Where the rank is refined (modular addition and IOI), ablations serve only to pick the cutoff rank $r$ along ACE's ordering (and, in IOI, the layer).
The resulting subspace is guaranteed neither to be compact nor to generalize:
that ACE's ordering passes both tests with a fraction of the dimensions random or reversed orderings need (\figpanel{fig:sup_modadd_prefix}{}),
and that IOI cores pass them on prompts not used to set their ranks while rank-matched PCA and random subspaces do not (\tabl{tab:ioi_heldout}; \figpanel{fig:ioi_wang_controls}{}),
is an empirical property of the learned representations.

\paragraph{When activity and relevance align.}
ACE contains PCA and contrastive activation steering as special cases.
When sensitivity is isotropic ($\mathbf{J}^\top\mathbf{J} \propto \mathbf{I}$), ACE recovers PCA;
for a binary task with a least-squares linear readout, rank-1 ACE recovers the difference of means $\boldsymbol{\mu}_1 - \boldsymbol{\mu}_0$ (derived below).
The PCA case is even expected for simple tasks,
when there is no inherent pressure for models to ``hide'' computations in low-variance subspaces.
In more complex models, however, high-variance directions are unlikely to cleanly align with target tasks.
Still, the distinction matters even when the subspaces agree:
PCA identifies where variance concentrates;
ACE identifies where the input--output map flows, and ablation tests whether that subspace is causally relevant.
In other words, PCA is descriptive and statistical, whereas a validated ACE core is also causal,
licensing downstream treatment and interpretation of the returned subspace
and its fitted operator as a dynamical system realizing a causal algorithm.

\paragraph{Rank-1 ACE and the difference of means.}
Let $\mathbf{y} \in \{0,1\}^N$ be the class labels, $\mathbf{H}$ mean-centered, and $\mathbf{w} = (\mathbf{H}^\top\mathbf{H})^{-1}\mathbf{H}^\top\mathbf{y}$ the ordinary least-squares output weight.
Every Jacobian equals $\mathbf{w}^\top$, so $\mathbf{J} = \mathbf{1}\mathbf{w}^\top$, and $\mathbf{H}\mathbf{J}^\top = (\mathbf{H}\mathbf{w})\mathbf{1}^\top$ has rank one with $\mathbf{U}_1 \propto \mathbf{H}\mathbf{w}$.
The covariance inverse then cancels in the ACE projection:
\begin{equation*}
\mathbf{q}_{\mathrm{ACE}} \propto \mathbf{H}^\top\mathbf{U}_1 \propto (\mathbf{H}^\top\mathbf{H})\mathbf{w} = \mathbf{H}^\top\mathbf{y} \propto \boldsymbol{\mu}_1 - \boldsymbol{\mu}_0.
\end{equation*}
Contrastive steering~\citep{turner2023activation,rimsky2024steering} is thus a special case of ACE, which explains their close agreement on subject--verb agreement ($|\cos| = 0.92$; \figpanel{fig:sup_sva_controls}{}).
Where the relevant computation is not aligned with the dominant variance, or needs several dimensions or sites, as in IOI, PCA and rank-1 directions fail where ACE does not (\hyperref[fig:sup_ioi_single]{Figs.~\ref*{fig:sup_ioi_single}} and~\ref{fig:ioi_wang_controls}):
ACE handles multi-dimensional cores, nonlinear readouts through the Jacobian, and multiple sites.

\paragraph{Rank bounds.}
For a linear readout target, the Jacobian rows lie in the readout span, so the core's rank is at most the number of independent readout directions.
For scalar margin tasks where the map from the hidden state to the logits is nonlinear,
each prompt contributes its own rank-1 Jacobian outer product, and the bound is the number of Jacobian samples, not the number of readout directions.

\section{Markov chain experiment}
\label{app:markov}
Three single-layer transformers
($d_{\mathrm{model}} = 64$, $d_{\mathrm{ff}} = 256$, $|V| = 4$)
with causal attention masking
were trained with independent random seeds on next-token prediction
for sequences generated by a four-state Markov chain.

The Markov chain transition probability matrix,
\begin{align*}
    \mathbf{T} \coloneq \begin{pmatrix}
        \alpha & \beta & 0 & 0 \\
        0 & \alpha & \beta & 0 \\
        0 & 0 & \alpha & \beta \\
        \beta & 0 & 0 & \alpha
    \end{pmatrix},
\end{align*}
was instantiated with $\alpha = 0.75$
and $\beta = 0.25$,
yielding eigenvalues (spectrum)
$\lambda \in \{1, 0.75 + 0.25\ii, 0.75 - 0.25\ii, 0.5\}$,
and has stationary distribution $\sta = \left[0.25, 0.25, 0.25, 0.25\right]$.

Training used \texttt{AdamW} with learning rate $10^{-3}$ and no weight decay
for 40 epochs on 3{,}000 sequences of length 32 generated by $\mar$, with batch size 64.

Trained model performance is compared against two baselines:
\begin{align*}
    \text{Chance:} \quad & \max (\sta), \\
    \text{Bayes-optimal:} \quad & \sum_i \pi_i \max_j T_{ij},
\end{align*}
Chance accuracy reflects always predicting the most common token;
Bayes-optimal accuracy reflects the best possible one-step prediction
given the stochastic nature of the chain.

Algorithmic cores were extracted using a 99.9\% rank energy
threshold without ablation-refinement,
$\mathbf{H}$ was computed for all test activations,
and
$\mathbf{J}$ was defined by the target function $f(\mathbf{h}) \coloneq \mathrm{logits}(\mathbf{h})$.

\paragraph{Fitting dynamics.}
Hidden state (mean-centered) sequences were projected into core coordinates
\(\mathbf{z}_t = \mathbf{Q}^\top \mathbf{h}_t\)
and a linear operator was fit by least squares to predict next-step dynamics,
\begin{equation*}
\mathbf{z}_{t+1} \approx \mathbf{A}\mathbf{z}_t.
\end{equation*}
The spectrum of $\mathbf{A}$ was used to characterize the learned dynamics.
When comparing fitted operators in the core to ground truth,
the Perron--Frobenius eigenvalue $\lambda = 1$ of $\mar$ (corresponding to the stationary distribution) is excluded,
as it reflects normalization.

To calibrate, core operator fits were compared against an \emph{oracle} ceiling for next-token prediction:
\begin{equation*}
R_{\rm oracle}^2 \coloneq 1-\frac{1}{|V|}\,\mathbf{1}^\top\!\left(\mathbf{m}\oslash \mathbf{v}\right),
\end{equation*}
where
\begin{equation*}
\mathbf{m}\coloneq \mathrm{diag}(\boldsymbol{\pi})\big(\mathbf{T}\odot(\mathbf{1}-\mathbf{T})\big)^\top \mathbf{1},
\qquad
\mathbf{v}\coloneq \boldsymbol{\pi}\odot(\boldsymbol{1}-\boldsymbol{\pi}),
\end{equation*}
and $\odot$ and $\oslash$ denote elementwise product and division.

\subsection*{Tables}

\begin{table}[H]
\centering
\caption{Transformer Markov-chain test accuracy: \texttt{Full Model}:
no ablations; \texttt{Core-only}: ablating non-core dimensions;
\texttt{Core-removed}: ablating core (Methods).
Data are plotted in \figpanel{fig:markov}{B}.}
\label{tab:mark_ablations}
\begin{tabular}{lccc}
\toprule
 & Full Model & Core-only & Core-removed \\
\midrule
M$_1$ & 0.748 & 0.748 & 0.261 \\
M$_2$ & 0.748 & 0.748 & 0.237 \\
M$_3$ & 0.748 & 0.748 & 0.247 \\
\bottomrule
\end{tabular}
\end{table}

\begin{table}[H]
\centering
\caption{Pairwise core geometry and CCA similarity.
Projector overlap is the squared Frobenius overlap between core subspaces;
angles are principal angles (degrees); CCA lists canonical correlations.
Overlap and mean CCA are visualized in \figpanel{fig:markov}{C,D}.}
\label{tab:mark_cca}
\begin{tabular}{lcccc}
\toprule
Pair & Proj. Overlap & Principal Angles & CCA \\
\midrule
M$_1$--M$_2$ & 0.027 &  [78.4, 79.6, 85.4] & [0.999, 0.999, 0.927] \\
M$_1$--M$_3$ & 0.031 & [76.4, 80.1, 85.2] & [0.999, 0.999, 0.949] \\
M$_2$--M$_3$ & 0.027 & [76.0, 81.5, 88.7] & [0.999, 0.999, 0.958] \\
\bottomrule
\end{tabular}
\end{table}

\begin{table}[H]
\centering
\caption{Eigenvalues $\left\{ \lambda_i \right\}$ from operators fit
in transformer cores compared with those from the Markov transition
probability matrix $\mar$ (excluding the Perron--Frobenius eigenvalue).
Spectral overlap is visualized in \figpanel{fig:markov}{E}.}
\label{tab:mark_spec}
\begin{tabular}{lcccc}
\toprule
& Markov chain & Core$_1$ & Core$_2$ & Core$_3$ \\
\midrule
$\lambda_1$ & $0.75 + 0.25\ii$ & $0.75 + 0.25\ii$  & $0.75 + 0.25\ii$ & $0.75+0.25\ii$ \\
$\lambda_2$ & $0.75 - 0.25\ii$ & $0.75 - 0.25\ii$ & $0.75 - 0.25\ii$ & $0.75-0.25\ii$ \\
$\lambda_3$ & $0.50$ & $0.51$ & $0.49$ & $0.48$ \\
\bottomrule
\end{tabular}
\end{table}

\subsection{Null controls and rank selection}
\label{app:markov_ladder}

\begin{figure}[H]
\centering
\includegraphics[width=\linewidth]{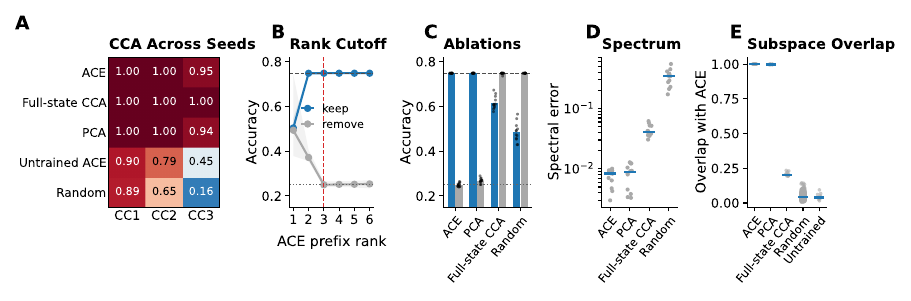}
\caption{\textbf{Markov null controls} (12 seeds). (A) Canonical correlations between rank-3 subspaces of different seeds
(66 pairs): ACE, full-state CCA-3, random, untrained-model ACE, and PCA. (B) Core-only (keep) and core-removed accuracy against
ACE prefix rank; the causal cutoff is at rank 3. (C) Keep (blue) and remove (gray) for rank-matched controls. (D) Normalized error of the
spectrum fitted in each subspace against the chain's spectrum. (E) Projector overlap with the ACE core.}
\label{fig:sup_markov}
\end{figure}

Twelve single-layer transformers were trained from independent seeds on the four-state chain above and compared with rank-matched controls (\figpanel{fig:sup_markov}{}).
The first two canonical correlations are high even for nulls (0.89 and 0.65 for random rank-3 subspaces; 0.90 and 0.79 for ACE on an untrained model);
the third separates them (0.95 for ACE, 0.16 for random subspaces, and 0.45 for untrained models),
and only ACE and PCA pass both ablations and recover the chain's spectrum (\figpanel{fig:sup_markov}{C,D}).
ACE and PCA find the same core on this task, as expected when activity and relevance align (Appendix~\ref{app:cores}; projector overlap $> 0.99$; \figpanel{fig:sup_markov}{E}).
A full-state CCA-3 alignment across seeds finds correlated directions that are neither sufficient nor necessary,
and removing a random rank-3 subspace leaves accuracy at baseline (0.75).

\begin{figure}[H]
\centering
\includegraphics[width=0.8\linewidth]{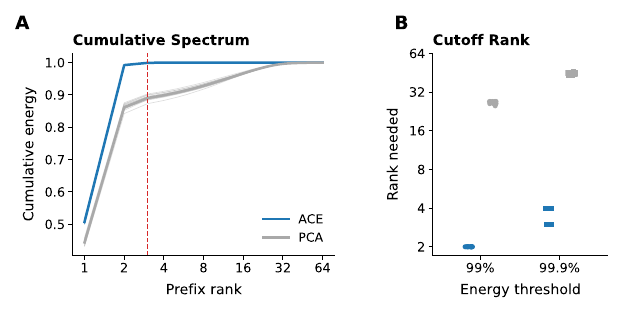}
\caption{\textbf{Markov energy.} (A) Cumulative energy against prefix rank: ACE balanced covariance$\times$sensitivity
energy and PCA activation variance (thin lines: models; thick: mean). (B) Rank needed to reach 99\% and 99.9\% of the
energy (12 seeds).}
\label{fig:sup_markov_energy}
\end{figure}

The causal cutoff coincides with the spectral criterion (\figpanel{fig:sup_markov}{B} and \figpanel{fig:sup_markov_energy}{}):
removing the first three ACE directions brings accuracy to chance in all 12 seeds (0.24--0.27; chance 0.25),
and those three directions hold over 99.8\% of ACE's balanced energy, against 89\% of the activation variance for the first three principal components.
Reaching 99.9\% takes 3--4 ACE directions and 43--46 principal components.

\section{Modular addition experiment}
\label{app:modadd}
Three two-layer transformers ($d_{\mathrm{model}} = 128$, $d_{\mathrm{ff}} = 512$, $|V| = 53$)
were trained on $a + b \equiv c \pmod{53}$.
The dataset consists of all $53^2 = 2809$ input pairs,
split evenly into train and test sets with a fixed random seed.
Input sequences are $[a, b]$ with target $[b, c]$.

Training used \texttt{AdamW} with learning rate $10^{-3}$, batch size 512, dropout $0.1$, and weight decay $\omega = 1$.
Models were trained for $2 \times 10^4$ epochs, with core extraction performed every 100 epochs.
The grokking epoch was defined as the first analysis time point at which all three models achieved perfect test accuracy,
which occurred at epoch 800.

To study the effect of continued weight decay after grokking,
at epoch 900, transformers were ``branched'' -- duplicated and split into two regimes -- where
weight decay was either maintained at $\omega = 1$
or disabled ($\omega \to 0$) for the remainder of training.
Dropout (0.1) stayed on in both branches, so the comparison isolates weight decay under matched dropout: the branch without weight decay was still perturbed and regularized, yet did not inflate.

For core extraction,
$\mathbf{H}$ was computed over all test-set activations,
and $\mathbf{J}$ was estimated using 64 Jacobian samples,
defined by the target function $f(\mathbf{h}) \coloneq \mathrm{logits}(\mathbf{h})$.
Core rank was selected first via the 99\% energy threshold, and then
refined with ablations to ensure causal importance.

For operator fitting, centroids $\bar{\mathbf{r}}_c$ were computed
as the centered mean core activation over all test examples with answer token $c$.
A linear shift operator $\mathbf{A}$ satisfying,
\begin{equation*}
\bar{\mathbf{r}}_{(c+1) \bmod 53} \approx \mathbf{A}\bar{\mathbf{r}}_c
\end{equation*}
was fit by ridge-regularized least squares after dimensionality reduction with SVD.
Generalization was evaluated by holding out cycle transitions rather than examples:
the 53 answer classes were split into disjoint calibrate/evaluate sets by selecting a contiguous block of classes
for the evaluate set, and the fit was performed only on transitions $c\!\to\!c{+}1$
whose endpoints both lie in the calibration class set;
evaluation used only transitions whose endpoints both lie in the evaluate class set and fit is denoted as $R_h^2$.
For descriptive fits, $R^2$ is reported without holding out transitions or ridge-regularization.

To summarize spectral structure, eigenvalues of $\mathbf{A}$ with magnitude close to 1 were identified as rotational modes,
and each such mode was assigned a frequency bin by rounding its angle to the nearest integer multiple of $2\pi/53$.
Because complex-conjugate eigenvalue pairs correspond to the same oscillation up to direction,
bins $k$ and $53-k$ were mapped to the same bin. This implies a maximum of
$\lfloor 53/2\rfloor + 1 = 27$ distinct bins: one $k=0$ bin and 26 nonzero oscillatory bins.
Mode count is defined as the number of occupied nonzero bins,
and derives from operators fit without holding out transitions,
since the goal is descriptive characterization rather than generalization evaluation.

\subsection{Core dimension, ordering, and replicate seeds}
\begin{figure}[H]
\centering
\includegraphics[width=\linewidth]{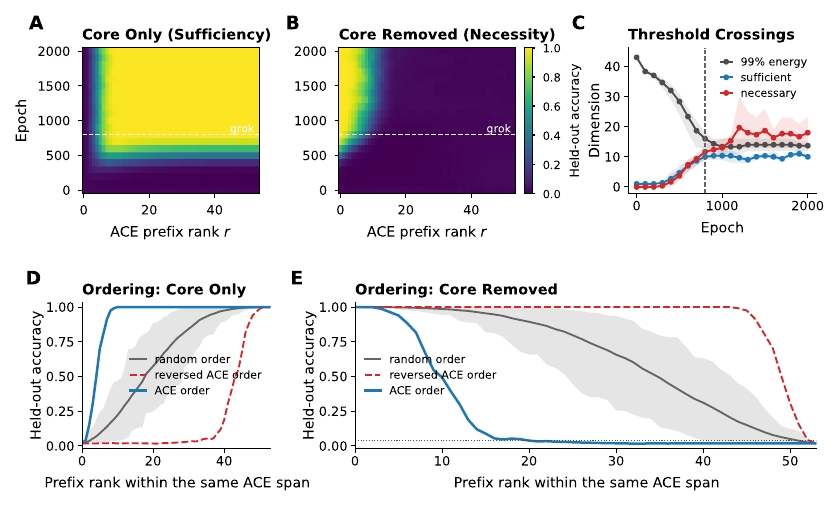}
\caption{\textbf{Modular addition: prefix-rank analysis} (weight decay $\omega = 1$, three models, epochs 0--2{,}000).
(A, B) Held-out accuracy with only the first $r$ ACE directions kept (A) or removed (B), against epoch. (C) Dimensions at
which the 99\% energy, sufficiency, and necessity thresholds are crossed; dotted: baseline accuracy. (D, E) Ordering control at
epoch 2{,}000: ACE order, 50 random orders, and reversed ACE order within the same span.}
\label{fig:sup_modadd_prefix}
\end{figure}

\paragraph{Sufficient and necessary dimensions.}
At each checkpoint, ablations were evaluated across prefix ranks $r$ of the ACE basis (\figpanel{fig:sup_modadd_prefix}{}).
The sufficient dimension is the smallest prefix whose core-only accuracy is within 0.001 of baseline,
and the necessary dimension is the smallest prefix whose removal brings accuracy to $2/|V|$ or below.
The operational core rank is the largest of the 99\%-energy rank and these two crossings.
With a linear readout on $|V| = 53$ tokens, the ACE spectrum has at most 53 nonzero values, so core dimensions above 53 (30\% of model-epochs after epoch 7{,}000; at most 68) count prefix directions beyond it.
Prefix sweeps were repeated at epoch 2{,}000 in 50 random orders and in reversed order within the same 53-dimensional ACE span.
Both take far more dimensions for both tests
(sufficiency at 44 and 51 dimensions on average, against 10 in ACE order; necessity at 49 and 53, against 18; \figpanel{fig:sup_modadd_prefix}{D,E}).

\begin{figure}[H]
\centering
\includegraphics[width=\linewidth]{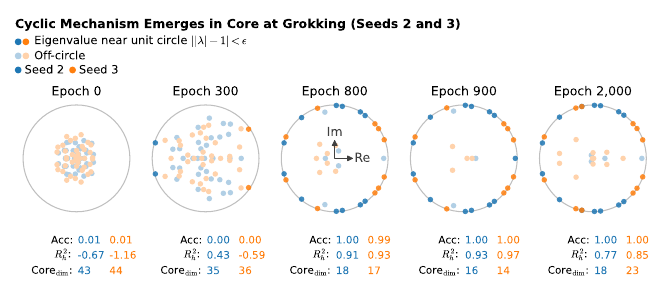}
\caption{\textbf{\figpanel{fig:mod1}{C} for the other two models.} Eigenvalues of the shift operator fitted in each core, for seed
2 (blue) and seed 3 (orange) on the same circles; pale: off the unit circle ($||\lambda| - 1| \ge \epsilon$,
$\epsilon = \sin(\pi/53)$). Same checkpoints, analysis, and held-out fit as the main panel, whose values this pipeline
reproduces for seed 1.}
\label{fig:sup_modadd_seeds}
\end{figure}

\section{Subject--verb agreement experiment}
\label{app:sva}
GPT-2 Small (124M parameters, 12 layers), Medium (355M, 24 layers),
Large (774M, 36 layers), LLaMA-3.1 (8B, 32 layers), Gemma-2 (9B, 42 layers), and Qwen2.5 (32B, 64 layers)
were analyzed on subject--verb number agreement.
All are base (not instruction-tuned) Hugging Face checkpoints: \texttt{gpt2}, \texttt{gpt2-medium}, \texttt{gpt2-large}, \texttt{meta-llama/Llama-3.1-8B}, \texttt{google/gemma-2-9b}, and \texttt{Qwen/Qwen2.5-32B}.

\paragraph{Prompts.}
A dataset of 1{,}200 prompts (600 singular, 600 plural) was constructed
by combining head nouns (for example, ``key''/``keys'', ``child''/``children'')
with attractor nouns of opposite number (e.g., ``cabinets''/``cabinet'')
via connectors (``to the'', ``near the'', ``next to the'', etc.).
Five syntactic templates were used:
base (``The key to the cabinets''),
front-padded (``In this ancient kingdom, the key to the cabinets''),
back-padded (``The key to the cabinets in the old kingdom''),
existential (``There key near the boxes''),
and relative clause (``The key that guards the cabinets'').
Half of prompts were prefixed with ``In the past,'' to vary tense context.
The dataset was split evenly into train and test sets.
Note: some prompts deliberately employ ungrammatical word order (\textit{e.g.}, ``There key near the boxes'') to
assess whether the agreement core remains robust to structural violations,
forcing the model to resolve agreement based on the head noun
rather than positional heuristics.

\begin{figure}[H]
\centering
\includegraphics[width=0.8\linewidth]{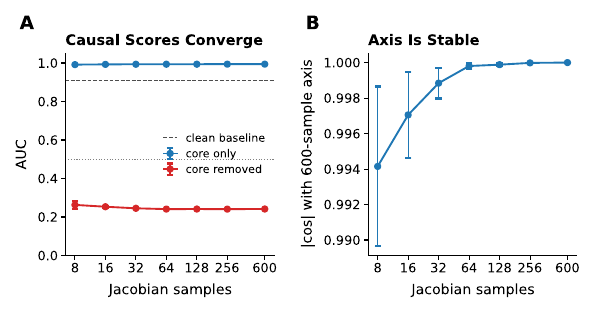}
\caption{\textbf{Subject--verb agreement Jacobian robustness} (GPT-2 Small, layer 11): axis alignment with the 600-sample axis and core-only and
core-removed AUC against the number of Jacobian samples.}
\label{fig:sup_jacobian}
\end{figure}

\paragraph{Target function.}
The number margin was defined on the final-token hidden state $\mathbf{h}$:
\begin{align*}
    f(\mathbf{h}) \coloneq (\mathrm{logit}_{\textit{are}} + \mathrm{logit}_{\textit{were}})
                  - (\mathrm{logit}_{\textit{is}} + \mathrm{logit}_{\textit{was}}).
\end{align*}
$\mathbf{J}$ was accumulated over all 600 extraction prompts (one Jacobian of $f$ per prompt, evaluated at the final token), though subsampling shows that the axis and its causal scores converge by 64 samples (\figpanel{fig:sup_jacobian}{}).

\paragraph{Layer sweep.}
Candidate cores were extracted at each layer and evaluated via ablation.
For each model, the layer with maximal flip effect was selected as the core location.

\paragraph{Adaptive generation steering.}
For open-ended generation, a per-token adaptive intervention was applied during autoregressive decoding.
Let $\mathbf{q} \in \mathbb{R}^D$ denote the (unit-norm) core axis
and $\boldsymbol{\mu}$ the mean activation at the intervention layer.
The intervention reflects the hidden state $\mathbf{h}$ at the last token position
through the hyperplane orthogonal to the core axis:
\begin{align*}
    \tilde{\mathbf{h}} = \mathbf{h} - 2s \left[ (\mathbf{h} - \boldsymbol{\mu})^\top \mathbf{q} \right] \mathbf{q},
\end{align*}
where $s$ is a per-token steering strength determined adaptively.

At each decoding step, three forward passes are performed.
First, a \emph{gating} check: a clean forward pass (with $s=0$) computes the
softmax probability mass on the agreement-relevant verb tokens
($\textit{is}$, $\textit{are}$, $\textit{was}$, $\textit{were}$).
If this mass falls below a threshold, the token is unlikely to involve
an agreement decision and no intervention is applied ($s^\ast = 0$).

Otherwise, the steering strength is calibrated to produce a minimal margin flip.
Define the \emph{generation margin} as
$m \coloneq \log\!\sum_{v \in \{\textit{are},\,\textit{were}\}} e^{\ell_v}
     - \log\!\sum_{v \in \{\textit{is},\,\textit{was}\}} e^{\ell_v}$,
where $\ell_v$ denotes the logit for token $v$.
This logsumexp margin more accurately reflects the probability-space
competition between singular and plural verb groups than the linear
logit sum used for core extraction, where operating-point independence
of the Jacobian is preferred.
The calibration proceeds as:
(1)~the current margin $m_0$ is measured under the clean pass;
(2)~a small probing perturbation at strength $s_0$
    estimates the local gain $g = (m_1 - m_0)/s_0$;
(3)~the intervention strength is set to
    $s^\ast = (m_{\mathrm{target}} - m_0) / g$,
    where $m_{\mathrm{target}} = -\operatorname{sign}(m_0)\,\varepsilon$
    targets the minimal margin crossing with buffer $\varepsilon$.
An optional cap $|s^\ast| \leq s_{\mathrm{cap}}$ prevents extreme extrapolation.
This adaptive approach produces grammatical inversions
while minimizing collateral disruption to non-agreement tokens.

\subsection*{Tables}
\begin{figure}[H]
\centering
\includegraphics[width=0.85\linewidth]{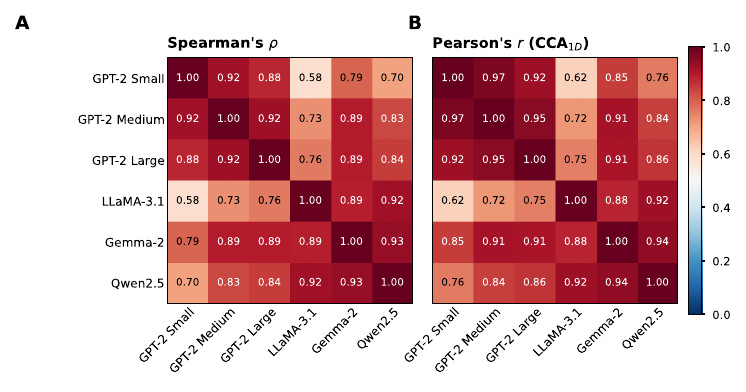}
\caption{Similarity of core coordinates (from \figpanel{fig:gpt2}{B}) across six language models: (A) Spearman's
$\rho$ (rank correlation) and (B) Pearson's $r$, whose magnitude equals CCA for one dimension.}
\label{fig:sva_cca}
\end{figure}

\begin{table}[H]
\centering
\caption{A one-dimensional subject--verb agreement core was extracted from each model,
despite massive variation in training, model parameterizations, and architectures (model dimension, number of layers).
Extracted core size ($d_{\rm core}$) is supported by the large \emph{spectral gap} (ratio of largest two singular value squares $\sigma_1^2/\sigma_2^2$).
The large spectral gaps indicate that these subspaces are effectively one-dimensional.
}
\label{tab:gpt2_sizes}
\begin{tabular}{lcccccc}
\toprule
Model & Parameters & Layers & $d_{\rm model}$ & $d_{\rm core}$ & Spectral gap & Core location (layer) \\
\midrule
GPT-2 Small & 124\,M & 12 & 768  & 1 & $40$ & 11 \\
GPT-2 Medium & 355\,M & 24 & 1024  & 1 & $44$ & 22 \\
GPT-2 Large & 774\,M & 36 & 1280   & 1 & $2.8 \times 10^{10}$ & 36 \\
LLaMA-3.1 & 8\,B & 32 & 4096 & 1 & $266$ & 28 \\
Gemma-2 & 9\,B & 42 & 3584 & 1 & $133$ & 42 \\
Qwen2.5 & 32\,B & 64 & 5120 & 1 & $531$ & 62 \\
\bottomrule
\end{tabular}
\end{table}

\begin{table}[H]
\centering
\caption{Agreement performance (AUC; 1 = perfect, 0.5 = chance, 0 = inverted) under core ablations across all six models.
Core-only preserves agreement (sufficiency),
core-removed collapses it below chance (necessity),
and core-flipped inverts grammatical number preferences (induces near perfect \emph{disagreement}).}
\label{tab:gpt2_ablations}
\begin{tabular}{lcccc}
\toprule
Model & Baseline & Core-only & Core-removed & Core-flipped \\
\midrule
GPT-2 Small  & 0.911 & 0.994 & 0.241 & 0.038 \\
GPT-2 Medium & 0.934 & 0.997 & 0.217 & 0.023 \\
GPT-2 Large  & 0.921 & 0.975 & 0.244 & 0.021 \\
LLaMA-3.1    & 0.808 & 0.918 & 0.209 & 0.092 \\
Gemma-2      & 0.886 & 0.978 & 0.102 & 0.035 \\
Qwen2.5      & 0.890 & 0.949 & 0.213 & 0.076 \\
\bottomrule
\end{tabular}
\end{table}

\subsection{Controls and robustness}
\label{app:sva_controls}

\paragraph{Scalar-margin Jacobians.}
Each prompt contributes its own Jacobian row, so a scalar margin does not force a one-dimensional core (Appendix~\ref{app:cores});
the second direction holds 1/531 to 1/40 of the first's energy (\tabl{tab:gpt2_sizes}).
The exception, GPT-2 Large, is selected at its final normalized state, where the map to the logits is linear, so its gap there is numerical;
one layer earlier, where the map is nonlinear, the ratio is still about 1/2{,}500 (on prompts where subject and attractor number vary independently).

\begin{figure}[H]
\centering
\includegraphics[width=0.45\linewidth]{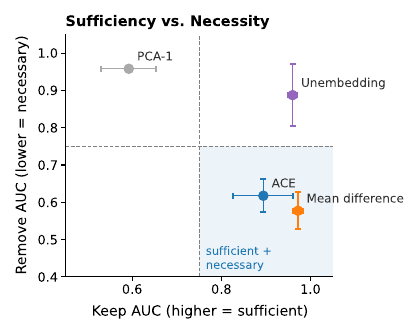}
\caption{\textbf{Subject--verb agreement: sufficiency against necessity} (mean $\pm$ 1 s.d. across six models; shading: keep $\ge 0.75$ and
remove $\le 0.75$). Same prompts as \figpanel{fig:sup_sva_controls}{}.}
\label{fig:sup_sva_keep_remove}
\end{figure}

\begin{figure}[H]
\centering
\includegraphics[width=\linewidth]{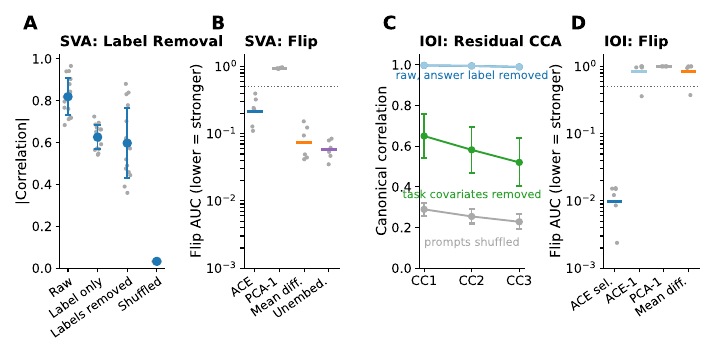}
\caption{\textbf{Language-model nulls.} (A) Subject--verb agreement cross-model correlation (15 pairs): raw, label-only expectation, labels
removed, prompts shuffled. (B) Subject--verb agreement flip AUC for ACE, PCA-1, difference of means, and unembedding (log scale; six models).
(C) IOI canonical correlations between single-site cores (15 pairs): raw, answer label removed, all task covariates removed,
prompts shuffled. (D) IOI flip AUC at a single site: selected-rank ACE, ACE rank 1, PCA-1, and difference of means.}
\label{fig:sup_nulls}
\end{figure}

\begin{table}[H]
\centering\small
\caption{\textbf{Subject--verb agreement by prompt class}: AUC on grammatical prompts and on the deliberately ungrammatical
existential (``There'') prompts.}
\label{tab:sup_sva_classes}
\begin{tabular}{lcc}
\toprule
Model & Grammatical & ``There'' (ungrammatical) \\
\midrule
GPT-2 Small & 0.920 & 0.973 \\
GPT-2 Medium & 0.970 & 0.965 \\
GPT-2 Large & 0.957 & 0.974 \\
LLaMA-3.1 & 0.783 & 0.684 \\
Gemma-2 & 0.899 & 0.766 \\
Qwen2.5 & 0.862 & 0.768 \\
\bottomrule
\end{tabular}
\end{table}

\section{Grokking dynamics}
\label{app:grok}
\subsection{Mathematical model}
\label{app:grok_math}
Let $\avec(t)\in\R^\mu$ denote the mode coefficients and let $\dvec\in\R^\mu$ be fixed with
$\psi_k \coloneq 1-\cos(2\pi k/p)$.
Define the (test-relevant) margin
$m(t)\coloneq \langle \avec(t), \dvec\rangle$.

Post-memorization, training loss is approximately zero.
Updates are driven by the weight decay penalty $-\omega\avec(t)$ and a minimal corrective motion $\gamma(t)\dvec$ needed to remain on the zero-loss manifold, plus zero-mean stochasticity $\xi(t)$ (optimizer noise).

\vspace{0.25em}
\textbf{Direction of $\dvec$.} Among all infinitesimal updates $\Delta\avec$ that increase the margin by one unit, the minimum-norm update solves
\begin{equation*}
\arg\min_{\Delta\avec}\|\Delta\avec\|_2 \quad \text{s.t.}\quad \langle \Delta\avec,\dvec\rangle = 1.
\end{equation*}
By the Cauchy--Schwarz inequality, the solution is $\Delta\avec = \dvec/\|\dvec\|_2^2$. Thus, the corrective gradient direction is strictly parallel to $\dvec$.

\vspace{0.25em}
\textbf{Margin dynamics.}
Differentiating $m(t)=\langle \avec(t),\dvec\rangle$ and isolating the noise-free deterministic trajectory yields the scalar ODE:
\begin{equation*}
\dot m(t) \;=\; -\omega\,m(t) \;+\; \gamma(t)\,\|\dvec\|_2^2.
\end{equation*}
Because weight decay is the only systematic drift pulling the network off the margin,
the mean corrective force is taken to scale proportionally to maintain zero loss: $\gamma(t)\approx c\,\omega$ for some constant $c$.

Assuming sufficient dimensional capacity ($p < d_{\mathrm{model}}$), the initial memorized state is unstructured, meaning it carries negligible margin ($m(0) \approx 0$). Substituting the corrective force yields a simple linear relaxation equation:
\begin{equation*}
\dot m(t)=-\omega m(t)+ c\,\omega\|\dvec\|_2^2,\qquad m(0) \approx 0.
\end{equation*}
Solving this ODE yields the exact continuous-time margin trajectory:
\begin{equation*}
m(t)=m^{\ast}\big(1-e^{-\omega t}\big),
\qquad \text{where} \quad
m^{\ast}= c\,\|\dvec\|_2^2 = \kappa p.
\end{equation*}

\vspace{0.25em}
\textbf{Predicting grokking time.}
Grokking occurs at the first-hitting time $\tau$ when the margin reaches the generalization threshold $\delta$. Solving $m(\tau) = \delta$ yields the continuous gradient-flow time:
\begin{equation*}
\tau(p) = -\frac{1}{\omega}\log\!\left(1-\frac{\delta}{\kappa p}\right).
\end{equation*}

To map this idealized ODE to discrete training steps, the physical constants are decoupled.
The scaling rate becomes:
\begin{equation*}
  \tau_{\mathrm{grok}}(p) = - \Omega \log\!\left(1-\frac{p_{\mathrm{crit}}}{p}\right), \qquad (p_{\mathrm{crit}} < p < d_{\mathrm{model}}).
\end{equation*}

Here, $p_{\mathrm{crit}} \coloneq \delta/\kappa$ is the \textit{architectural constant}, defining the absolute capacity floor limit independent of the optimizer.
Conversely, $\Omega \propto (\eta\omega)^{-1}$ is the \textit{optimizer constant}, an empirical parameter that captures the characteristic relaxation time while absorbing discrete step-size dynamics, learning rate, momentum, and adaptive preconditioning from the \texttt{AdamW} optimizer.

\textbf{Gradient flow and \texttt{AdamW}.} The ODE is written for continuous-time gradient flow on an $L_2$ penalty, whereas the networks were trained in discrete steps with \texttt{AdamW}.
\texttt{AdamW} applies weight decay outside its preconditioner, $\Delta\theta_t = -\eta\,(\hat m_t/(\sqrt{\hat v_t}+\epsilon) + \omega\,\theta_{t-1})$.
After memorization the task gradient has no systematic component, so the preconditioned term has zero mean and the expected drift is $\dot\theta = -\eta\omega\theta$, the $-\omega m(t)$ term of the margin equation;
the preconditioner's effect on the timescale is absorbed by the fitted constant $\Omega$.

\paragraph{Sweeps and fits.}
\label{app:grok_sweep}
One- and two-layer transformers ($d_{\mathrm{model}}=128$, $d_{\mathrm{ff}}=512$, \texttt{AdamW}, learning rate $10^{-3}$) were trained on a 50/50 train/test split of the $p^2$ input pairs, with 12 seeds per setting;
$\tau_{\mathrm{mem}}$ and $\tau_{\mathrm{gen}}$ are the first steps at which train and test accuracy reach 0.99, evaluated every step.
The weight-decay sweep fixed $p = 53$ and varied $\omega\in\{0.3, 0.5, 1, 1.5, 2, 3\}$ with minibatches of 512;
the modulus sweep fixed $\omega = 1$ and varied $p\in\{31, 43, 53, 61, 79, 89, 101\}$; $p = 17$, 13, and 11 (three seeds each), below the fitted $\hat p_{\mathrm{crit}}$, did not grok within 50{,}000 steps.
Because the dataset size grows quadratically with $p$, this sweep utilized full-batch gradient descent to ensure the empirical hitting time was isolated from dataset-dependent minibatch noise.
Exponents were fit by least squares in log--log space, and $\Omega$ and $p_{\mathrm{crit}}$ by nonlinear least squares on $\tau_{\mathrm{grok}}(p)$ above.

\subsection{One-layer sweeps and a control on the modulus sweep}
\label{app:grok_controls}

\begin{figure}[H]
\centering
\includegraphics[width=0.72\linewidth]{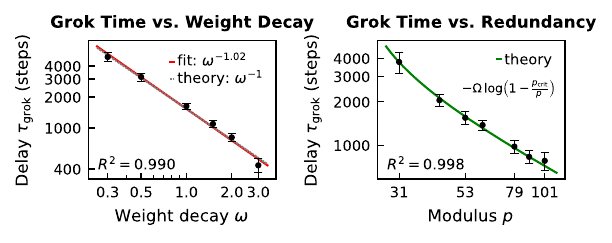}
\caption{\textbf{One-layer sweeps.} Delay $\tau_{\mathrm{grok}}$ against weight decay $\omega$ at $p = 53$ (left) and
against the modulus $p$ at $\omega = 1$ (right); mean $\pm$ s.d. over 12 seeds.}
\label{fig:grok_one_layer}
\end{figure}

\paragraph{One-layer sweeps.}
Repeating the weight-decay and modulus sweeps in one-layer transformers (same width, optimizer, and thresholds; 12 seeds per setting) gives quantitative agreement with the theory (\figpanel{fig:grok_one_layer}{}):
the delay scales with weight decay as $\tau_{\mathrm{grok}} \propto \omega^{-1.02}$ ($R^2 = 0.990$ at $p = 53$)
and follows the law over $p = 31$--101 with $\Omega = 2{,}767$ and $\pcrit = 23.1$ ($R^2 = 0.998$).

\begin{figure}[H]
\centering
\includegraphics[width=0.36\linewidth]{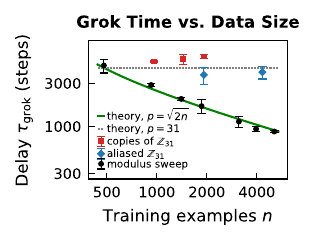}
\caption{\textbf{Control on the modulus sweep} (two-layer, mean $\pm$ s.d. over 3 seeds). $\tau_{\mathrm{grok}}$
against training examples. The modulus sweep follows the theory at $p = \sqrt{2n}$; copies and
aliased versions of $\mathbb{Z}_{31}$ stay near the theory at $p = 31$.}
\label{fig:grok_controls}
\end{figure}

\paragraph{Data size.}
With a 50/50 split, the training set grows as $p^2/2$, so a larger modulus also means more training examples.
To test whether the delay is set by the sample count $n$ rather than by the modulus, $n$ was varied at a fixed group, $\mathbb{Z}_{31}$, with the settings of the modulus sweep, in two ways (three seeds per setting, up to 30{,}000 steps; \figpanel{fig:grok_controls}{}):
\begin{itemize}\itemsep0pt
\item \emph{Copies.} One network was trained on $k = 2$--4 disjoint copies of $\mathbb{Z}_{31}$, with separate tokens and each copy split 50/50 on its own.
This grows the training set to $480k$ examples, and each copy is new arithmetic to learn.
\item \emph{Aliases.} Each number was given $a = 2$ or 3 interchangeable token names, with the split made over number pairs, so that every name combination of a pair falls on the same side.
This grows the training set to $480a^2$ examples with no new arithmetic.
\end{itemize}
If $n$ set $\tau_{\mathrm{grok}}$, both arms would fall to the theory's value at $p = \sqrt{2n}$, the modulus the sweep uses for $n$ examples.
Instead, every arm stays near the theory's value for $p = 31$:
copies run 18--34\% above it and aliases 10--16\% below, while the data-size prediction misses them by a factor of 2--4.
The learned solutions are comparable:
the embedding holds 90\% of its Fourier power in 5--10 frequencies per copy and 9--11 per aliased number, against 9--10 for a single $\mathbb{Z}_{31}$.

\section{Indirect object identification experiment}
\label{app:ioi}
GPT-2 Small, Medium, and Large, LLaMA-3.1, Gemma-2, and Qwen2.5 were analyzed on indirect object identification (IOI), using the base checkpoints of Appendix~\ref{app:sva}.

\paragraph{Prompts.}
Each stimulus was a pair of prompts that share a sentence frame, two single-token names ($n_1$ and $n_2$), and token positions,
and differ only in which name is repeated at the second mention ({\Stwo}), which flips the correct answer
(for example, ``When Mary and John went to the store, John gave a drink to'', and the same with ``Mary gave'').
Every prompt began with a start token.
Cores were extracted, and their ranks set, on 165 natural frames (the \emph{extraction set}).
They were tested on 27 held-out frames with 40 new names (the \emph{test set}; 1{,}026 prompts)
and on the 15 templates of Wang et al.~\citep{wang2022interpretability} (1{,}020 prompts).

\paragraph{Target function.}
The margin was defined at the final token ({\Eend}) as
\begin{align*}
    m \coloneq \mathrm{logit}_{n_1} - \mathrm{logit}_{n_2}.
\end{align*}

\paragraph{Core extraction.}
Cores were extracted at {\Stwo} and {\Eend} over 12{,}000 pairs,
with uncentered within-pair state differences in place of $\mathbf{H}$ and margin gradients in place of $\mathbf{J}$.
Layer $l$ denotes the state after block $l$.
As in subject--verb agreement, interventions act relative to the mean state $\boldsymbol{\mu}$: keeping the core gives $\tilde{\mathbf{h}} = \boldsymbol{\mu} + \mathbf{P}(\mathbf{h} - \boldsymbol{\mu})$, removing it gives $\tilde{\mathbf{h}} = \mathbf{h} - \mathbf{P}(\mathbf{h} - \boldsymbol{\mu})$, and flipping reflects the state through the core:
\begin{align*}
    \tilde{\mathbf{h}} = \mathbf{h} - 2\mathbf{P}(\mathbf{h} - \boldsymbol{\mu}).
\end{align*}

\begin{figure}[H]
\centering
\includegraphics[width=\linewidth]{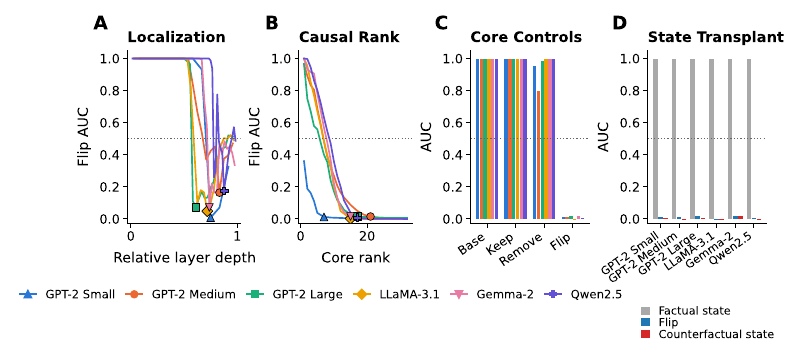}
\caption{\textbf{IOI single-site core} (one layer, final token; six models). (A) Flip AUC of a rank-12 core against relative depth; markers: selected layers. (B) Flip AUC against core rank at the selected layer; markers: selected ranks. (C) Baseline, keep, remove, and flip AUC at the selected core. (D) AUC after writing the prompt's own state into the core (factual, control), reflecting the core about the mean (flip), or writing the state a fitted affine encoder predicts for the counterfactual prompt (counterfactual).}
\label{fig:ioi_single_core}
\end{figure}

\begin{figure}[H]
\centering
\includegraphics[width=\linewidth]{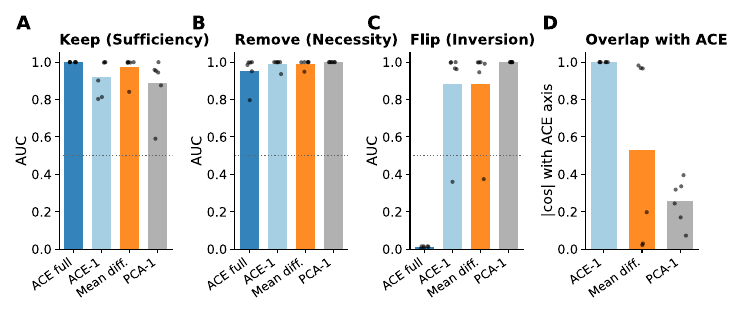}
\caption{\textbf{IOI single-site core: control and axis geometry} (six models; bars: mean; points: models). Keep, remove, and
flip AUC for the selected-rank ACE core, ACE rank 1, the difference of means, and PCA-1, and $|\cos|$ of the rank-1
directions with ACE rank 1.}
\label{fig:sup_ioi_single}
\end{figure}

\paragraph{Single-site cores.}
In all six models, a core was also extracted at {\Eend} alone, at the layer with maximal flip effect and the rank where flip AUC reaches its floor,
on a fixed set of templates with independent prompts and the tokenizer's default start token, present only in LLaMA-3.1 and Gemma-2 (\tabl{tab:review_single}, \hyperref[fig:ioi_single_core]{Figs.~\ref*{fig:ioi_single_core}} and~\ref{fig:sup_ioi_single}).
Its causal ranks (7, 21, 17, 15, 15, and 17) are at or below the 99\%-energy ranks (8, 24, 17, 17, 16, and 25).
The selected-rank core inverts the answer in every model (flip AUC $\approx 0$), while PCA-1 never moves it (1),
and ACE rank 1 and the difference of means invert it only partly in GPT-2 Small (0.36, 0.37) and not in the other models ($> 0.94$).
After residualizing all task variables and template identity, the leading canonical correlations across the 15 model pairs are 0.65, 0.58, and 0.52 (per pair 0.47--0.82),
against 0.29, 0.25, and 0.23 with prompts shuffled (\figpanel{fig:sup_nulls}{C}).

\paragraph{Layer sweep.}
Candidate cores were extracted at each layer at both sites (every other layer in Qwen2.5) and flipped at rank $\max(16,\ 2.5\%\text{ of }D)$.
Where flip AUC was at most 0.10, the layer with the lowest core-removed AUC at a fixed size, with the other site at a reference layer, was taken as the best layer at that site.

\begin{figure}[H]
\centering
\includegraphics[width=\linewidth]{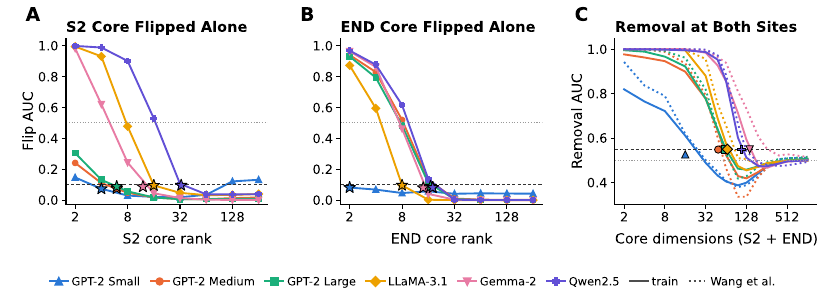}
\caption{\textbf{IOI rank selection.} (A, B) Each site's core flipped alone against its rank (dashed: the 0.10 cutoff; stars: sufficient rank). (C) Removal at both sites against core size at equal ranks, on the extraction set (solid) and the
prompts of Wang et al. (dotted). Dashed: the 0.55 cutoff; markers: the smallest core, on the extraction set.}
\label{fig:ioi_curves}
\end{figure}

\paragraph{Rank selection.}
The spectra have no gap at these ranks (ratio of successive singular values 1.01--1.08, and 1.46 at GPT-2 Small's {\Eend} core), so ranks were set by causal cutoffs (\figpanel{fig:ioi_curves}{}).
The \emph{sufficient core} is, at each site, the smallest rank that preserves the behavior when kept (AUC within 0.05 of clean) and inverts it when flipped (AUC at most 0.10).
The \emph{smallest core} ($r_{\text{\Stwo}} + r_{\text{\Eend}}$) is the smallest total that preserves the behavior when both sites are kept, is necessary when both are removed (AUC at most 0.55),
and inverts the answer when either site is flipped alone.
The sweep's best pair of layers and the reference pair were compared, and the pair with the smallest core was selected (\tabl{tab:ioi_cores}).

\begin{figure}[H]
\centering
\includegraphics[width=0.45\linewidth]{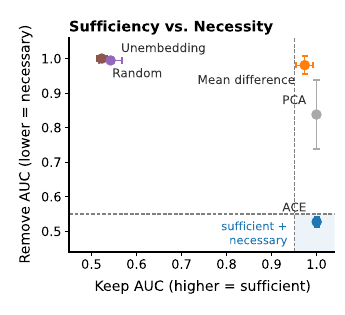}
\caption{\textbf{IOI: sufficiency against necessity} (mean $\pm$ 1 s.d. across GPT-2 Small, Medium, and Large; prompts of Wang et al.). Keep and remove at both sites. ACE, PCA of the activations, and a random subspace at each core's ranks; the difference of means and each prompt's unembedding direction at rank 1 per site. Shading: keep at least 0.95 and remove at most 0.55.}
\label{fig:ioi_wang_controls}
\end{figure}

\paragraph{Two-site results.}
At the ranks of the smallest core, on the extraction set, removing the {\Stwo} core alone leaves AUC at 0.76--0.88 and removing the {\Eend} core alone at 0.72--0.97 in five models, against at most 0.55 with both;
in GPT-2 Small, the {\Stwo} core alone reaches the cutoff (0.52), and the {\Eend} core alone gives 0.83.
At larger ranks, one site alone reaches the cutoff in some models: {\Stwo} at rank 13, 64, 448, and 200 in GPT-2 Small, GPT-2 Medium, Gemma-2, and Qwen2.5, and {\Eend} at rank 70 in GPT-2 Medium, but neither site in GPT-2 Large or LLaMA-3.1.
One-site cores do not reach the cutoff on the held-out prompts (0.58--0.69 on test, 0.59--0.66 on the Wang et al.\ prompts).
With equal ranks on the Wang et al.\ prompts (\figpanel{fig:ioi_curves}{C}), core-removed AUC is 0.51 in LLaMA-3.1 at 48 + 48, 0.54 in Qwen2.5 at 64 + 64, and 0.55 in Gemma-2 at 128 + 128.
On those prompts in the three GPT-2 models, only ACE is both sufficient and necessary (\figpanel{fig:ioi_wang_controls}{}):
core-removed AUC is 0.52--0.55 for ACE, 0.70--0.92 for PCA, 0.94--1 for the difference of means, 0.98--1 for each prompt's unembedding direction, and 1 for a random subspace.

\paragraph{Reduced model.}
Core coordinates are $\mathbf{z}_S \coloneq \mathbf{Q}_S^\top(\mathbf{h}_S - \boldsymbol{\mu}_S)$ and $\mathbf{z}_E \coloneq \mathbf{Q}_E^\top(\mathbf{h}_E - \boldsymbol{\mu}_E)$ at the smallest core,
with $n_1$ the name mentioned first.
The operators of Eq.~\eqref{eq:ioi_main} were fit by ordinary least squares on natural activations from the extraction set (4{,}000 pairs):
$\mathbf{A}$ on within-pair differences, $\mathbf{r}_E = \mathbf{A}\,\mathbf{r}_S$ with $\mathbf{r} \coloneq \tfrac12(\mathbf{z} - \mathbf{z}')$, and $\mathbf{c}$ and $d$ on single prompts.
No fit used an intervention.
On the test set, the {\Stwo} core alone gives the right answer (AUC $> 0.99$) and matches the network's choice on 95--100\% of prompts;
one direction at {\Stwo}, the mean within-pair difference, gives the same AUC as a readout but does not reliably control the answer ({\Stwo} mark, below).
$\mathbf{A}$ explains 41--90\% of the within-pair difference in the {\Eend} core and 61--87\% of its change when the {\Stwo} core state is replaced with the partner's (an {\Stwo} swap),
where it removes 55--78\% of the error left by the best single fixed direction (\tabl{tab:ioi_reduced}).

\paragraph{{\Stwo} mark.}
Let $\Delta\mathbf{c}$ be the difference between the two names in {\Stwo} core coordinates, each name read alone after the start token at the {\Stwo} layer,
and let $s = \pm 1$, opposite on the two prompts of a pair.
With $\mathbf{v}$ the extraction-set mean of $s(\mathbf{z}_S' - \mathbf{z}_S) - \Delta\mathbf{c}$, where $\mathbf{z}_S'$ is the partner's state, the intervention is
\begin{align*}
    \tilde{\mathbf{z}}_S = \mathbf{z}_S + s(\mathbf{v} + \Delta\mathbf{c}).
\end{align*}
It reverses the answer in all six models (AUC $< 0.03$), whereas neither $s\mathbf{v}$ nor $s\Delta\mathbf{c}$ alone reliably does so (0.45--1 and 0.31--0.98).

\paragraph{Transfer across models.}
$\mathbf{A}$ was carried from one model into another through linear maps between their cores,
fit on natural within-pair differences on half of the 652 prompt pairs shared by all six models, and scored on the other half (\tabl{tab:ioi_reduced}).
Over the 30 directed pairs, a carried $\mathbf{A}$ explains 60\% of the change in the target's {\Eend} core after an {\Stwo} swap,
against 77\% for the target's own and about 0\% with shuffled alignment,
and it removes 49\% of the error left by one fixed direction (69\% for the target's own).

\subsection*{Tables}
\begin{table}[H]
\centering
\small
\caption{Single-site IOI core ({\Eend} token; six models). Rank: causal rank. Core-only and core-removed: AUC with only the core kept or with the core removed. Flipped: AUC after flipping the core (ideal 0).
Transplant: AUC after writing into the core the state that an affine encoder, fit from the task variables to core coordinates, predicts for the counterfactual prompt (ideal 0).}
\label{tab:review_single}
\begin{tabular}{lcccccc}
\toprule
Model & Layer (depth) & Rank & Core-only & Core-removed & Flipped & Transplant \\
\midrule
GPT-2 Small & 9/12 (0.75) & 7 & 1.000 & 0.951 & 0.012 & 0.003 \\
GPT-2 Medium & 20/24 (0.83) & 21 & 1.000 & 0.797 & 0.015 & 0.001 \\
GPT-2 Large & 22/36 (0.61) & 17 & 1.000 & 0.985 & 0.015 & 0.006 \\
LLaMA-3.1 & 23/32 (0.72) & 15 & 1.000 & 0.998 & 0.002 & 0.002 \\
Gemma-2 & 31/42 (0.74) & 15 & 1.000 & 1.000 & 0.015 & 0.019 \\
Qwen2.5 & 56/64 (0.88) & 17 & 1.000 & 0.998 & 0.009 & 0.002 \\
\bottomrule
\end{tabular}
\end{table}

\begin{table}[H]
\centering
\small
\caption{IOI cores in six models, on the extraction set, where their ranks were set. Layers: {\Stwo} and {\Eend} sites (relative depth). Sufficient: at each site, the smallest rank that alone preserves and flips the behavior. Smallest core: the smallest $r_{\text{\Stwo}} + r_{\text{\Eend}}$ that also passes removal. AUC at the smallest core: core-only and core-removed at both sites, and flip at each site alone.}
\label{tab:ioi_cores}
\begin{adjustbox}{max width=\linewidth}
\begin{tabular}{lrccccccc}
\toprule
Model & $D$ & Layers {\Stwo}/{\Eend} & Sufficient & Smallest core (\% of $D$) & Core-only & Core-removed & Flip {\Stwo} & Flip {\Eend} \\
\midrule
GPT-2 Small & 768 & 6/9 (0.50/0.75) & 4 + 2 & 14 + 2 (2.1\%) & 0.986 & 0.526 & 0.022 & 0.081 \\
GPT-2 Medium & 1024 & 12/20 (0.50/0.83) & 6 + 18 & 12 + 37 (4.8\%) & 0.991 & 0.548 & 0.025 & 0.006 \\
GPT-2 Large & 1280 & 17/27 (0.47/0.75) & 6 + 17 & 24 + 38 (4.8\%) & 0.999 & 0.549 & 0.009 & 0.003 \\
LLaMA-3.1 & 4096 & 8/23 (0.25/0.72) & 16 + 8 & 32 + 35 (1.6\%) & 1.000 & 0.550 & 0.047 & 0.000 \\
Gemma-2 & 3584 & 13/33 (0.31/0.79) & 12 + 14 & 64 + 79 (4.0\%) & 1.000 & 0.549 & 0.003 & 0.000 \\
Qwen2.5 & 5120 & 24/56 (0.38/0.88) & 33 + 18 & 60 + 48 (2.1\%) & 1.000 & 0.549 & 0.038 & 0.001 \\
\bottomrule
\end{tabular}
\end{adjustbox}
\end{table}

\begin{table}[H]
\centering
\small
\caption{The cores of \tabl{tab:ioi_cores} on prompts not used to extract them or to set their ranks (AUC). Test: 27 held-out frames and 40 new names (1,026 prompts; clean $> 0.99$). Wang et al.: their 15 templates, 99 names, places, and objects (1,020 prompts; clean 1).}
\label{tab:ioi_heldout}
\begin{adjustbox}{max width=\linewidth}
\begin{tabular}{l|cccc|cccc}
\toprule
 & \multicolumn{4}{c|}{Test} & \multicolumn{4}{c}{Wang et al.} \\ Model & Core-only & Core-removed & Flip {\Stwo} & Flip {\Eend} & Core-only & Core-removed & Flip {\Stwo} & Flip {\Eend} \\
\midrule
GPT-2 Small & 0.998 & 0.608 & 0.006 & 0.064 & 1.000 & 0.547 & 0.001 & 0.037 \\
GPT-2 Medium & 0.998 & 0.529 & 0.027 & 0.004 & 1.000 & 0.517 & 0.003 & 0.001 \\
GPT-2 Large & 1.000 & 0.600 & 0.004 & 0.002 & 1.000 & 0.518 & 0.001 & 0.000 \\
LLaMA-3.1 & 1.000 & 0.647 & 0.048 & 0.000 & 1.000 & 0.633 & 0.071 & 0.000 \\
Gemma-2 & 1.000 & 0.691 & 0.006 & 0.004 & 1.000 & 0.706 & 0.001 & 0.000 \\
Qwen2.5 & 1.000 & 0.679 & 0.026 & 0.001 & 1.000 & 0.601 & 0.059 & 0.000 \\
\bottomrule
\end{tabular}
\end{adjustbox}
\end{table}

\begin{table}[H]
\centering
\small
\caption{Controls at the ranks of the smallest core, on the extraction set (AUC). PCA: principal components of the activations at each site, unpaired, over the same prompts as the core, mean subtracted. Random: a random subspace of the same size. Core-only and core-removed at both sites, and flip at each site alone.}
\label{tab:ioi_controls}
\begin{adjustbox}{max width=\linewidth}
\begin{tabular}{lc|ccc|ccc|ccc|ccc}
\toprule
 & & \multicolumn{3}{c|}{Core-only} & \multicolumn{3}{c|}{Core-removed} & \multicolumn{3}{c|}{Flip {\Stwo}} & \multicolumn{3}{c}{Flip {\Eend}} \\ Model & Ranks & ACE & PCA & Rand. & ACE & PCA & Rand. & ACE & PCA & Rand. & ACE & PCA & Rand. \\
\midrule
GPT-2 Small & 14 + 2 & 0.99 & 0.98 & 0.51 & 0.53 & 0.64 & 0.98 & 0.02 & 0.03 & 0.98 & 0.08 & 0.15 & 0.98 \\
GPT-2 Medium & 12 + 37 & 0.99 & 0.97 & 0.51 & 0.55 & 0.87 & 1.00 & 0.03 & 0.06 & 1.00 & 0.01 & 0.24 & 1.00 \\
GPT-2 Large & 24 + 38 & 1.00 & 0.99 & 0.52 & 0.55 & 0.82 & 1.00 & 0.01 & 0.03 & 1.00 & 0.00 & 0.29 & 1.00 \\
LLaMA-3.1 & 32 + 35 & 1.00 & 1.00 & 0.50 & 0.55 & 1.00 & 1.00 & 0.05 & 0.33 & 1.00 & 0.00 & 0.49 & 1.00 \\
Gemma-2 & 64 + 79 & 1.00 & 1.00 & 0.50 & 0.55 & 0.94 & 1.00 & 0.00 & 0.01 & 1.00 & 0.00 & 0.29 & 1.00 \\
Qwen2.5 & 60 + 48 & 1.00 & 1.00 & 0.51 & 0.55 & 0.90 & 1.00 & 0.04 & 0.07 & 1.00 & 0.00 & 0.63 & 1.00 \\
\bottomrule
\end{tabular}
\end{adjustbox}
\end{table}

\begin{table}[H]
\centering
\small
\caption{Share of the change in the {\Eend} core after an {\Stwo} swap explained by $\mathbf{A}$. Own: the model's $\mathbf{A}$, on the test set, with $\mathbf{A}$ fit on shuffled pairs as control.
Carried: $\mathbf{A}$ from the other five models (mean), on the shared pairs, with shuffled alignment as control; on those pairs, the model's own $\mathbf{A}$ is within 0.01 of the first row.}
\label{tab:ioi_reduced}
\begin{tabular}{lcccccc}
\toprule
 & \multicolumn{3}{c}{GPT-2} & & & \\
\cmidrule(lr){2-4}
 & Small & Medium & Large & LLaMA-3.1 & Gemma-2 & Qwen2.5 \\
\midrule
Own $\mathbf{A}$ & 0.87 & 0.61 & 0.74 & 0.81 & 0.83 & 0.79 \\
\quad control: shuffled pairs & 0.02 & $-0.01$ & $-0.01$ & 0.01 & 0.00 & $-0.02$ \\
Carried $\mathbf{A}$ & 0.89 & 0.50 & 0.50 & 0.63 & 0.59 & 0.51 \\
\quad control: shuffled alignment & 0.01 & $-0.01$ & 0.00 & $-0.01$ & $-0.01$ & $-0.02$ \\
\bottomrule
\end{tabular}
\end{table}

\section{Compute resources}
The Markov-chain and modular-addition analyses can be run on a laptop and completed without specialized compute. Local runs used a MacBook Pro (Apple M1 Max, 64 GB). The subject--verb agreement analyses can be run on either a MacBook or GPUs in Google Colab; the IOI extractions were run on Google Colab GPUs (A100 with 80 GB and G4 runtimes).

\section{Broader impacts}
\label{app:impacts}
Isolating a compact subspace that controls a behavior provides both a lever for altering it and a means of auditing it.
Flipping the agreement core induces systematic grammatical errors throughout open-ended generation (\figpanel{fig:sva_gen}{}),
and the same procedure could be aimed at behaviors that matter more: inducing errors that are hard to detect, or weakening safeguards that a model implements in a low-dimensional subspace.
The method needs white-box access (activations and gradients), so it does not apply to models served only through an API, but it is cheap for any open-weights model.
Conversely, a core that is necessary and sufficient for a behavior can be monitored,
and its removal or flipping is a direct test of whether a safeguard is carried robustly or by a single, easily disabled direction.

\end{document}